\documentclass{article}

\usepackage[T1]{fontenc}
\usepackage[utf8]{inputenc}
\usepackage{xcolor}
\usepackage{xparse}

\usepackage{geometry}
\usepackage{setspace}
\usepackage{float}
\usepackage{authblk}
\usepackage{lineno}

\usepackage{amsmath,amssymb,amsfonts,amsthm,bbold}
\usepackage{mathtools}
\usepackage{physics}
\usepackage{interval}
\usepackage{leftindex,tensor}
\usepackage[version=4]{mhchem}

\usepackage{graphicx}
\usepackage{caption}
\usepackage[labelformat=empty,skip=0pt]{subcaption}
\usepackage{tikz}
\usetikzlibrary{positioning, arrows.meta}
\usepackage{mdframed}
\usepackage{rotating}
\usepackage{tabularx}
\usepackage{booktabs}
\usepackage[most]{tcolorbox}

\usepackage{algorithm}
\usepackage{algpseudocode}

\usepackage[dvipsnames, svgnames]{xcolor}
\usepackage{xurl}
\usepackage[colorlinks=true,breaklinks=true,
            urlcolor=Blue,linkcolor=Blue,citecolor=Green]{hyperref}
\usepackage{cleveref}
\usepackage[backend=bibtex,bibencoding=utf8,language=auto,
            style=numeric-comp,sorting=none,maxbibnames=10,natbib=true]{biblatex}
\newcommand{\subfiglabel}[1]{\textbf{(\lowercase{#1})}}
\newcommand{\captiontitle}[1]{\textbf{#1}}
\crefname{figure}{Figure}{Figures}
\Crefname{figure}{Figure}{Figures}
\crefname{equation}{Equation}{Equations}
\Crefname{equation}{Equation}{Equations}
\crefname{section}{Section}{Sections}
\Crefname{section}{Section}{Sections}
\crefname{subsection}{Subsection}{Subsections}
\Crefname{subsection}{Subsection}{Subsections}
\crefname{table}{Table}{Tables}
\Crefname{table}{Table}{Tables}
\crefname{notation}{Notation}{Notations}
\Crefname{notation}{Notation}{Notations}
\creflabelformat{equation}{#2\textup{#1}#3}

\newcommand*\phantomsubfigure[1]{%
  \begin{subfigure}[]{0pt}\caption{}\label{#1}\end{subfigure}%
}

\usepackage{siunitx}
\DeclareSIUnit{\coreh}{core\cdot h}
\DeclareSIUnit{\spike}{spikes}
\DeclareSIUnit{\spikepersecond}{\spike\per\second}

\newcommand{\loss}{\mathcal{L}}
\newcommand{\filter}[3]{\mathcal{#1}_{#2}\qty({#3})}
\newcommand{\clash}[2]{#1^\qty(#2)}
\newcommand{\code}[1]{{\texttt{\detokenize{#1}}}}
\newcommand{\epropplus}{e-prop\textsuperscript{+}}
\newcommand{\dvr}[3][]{\frac{{\mathrm{d}}^{#1} #2}{{\mathrm{d}}^{#1} #3}}

\theoremstyle{definition}
\newtheorem{notation}{Notation}

\newcommand{\orcid}[1]{\href{https://orcid.org/#1}{ORCID: #1}}

\newcounter{suppfigure}

\newenvironment{suppfigure}[1][]{%
  \refstepcounter{suppfigure}%
  \begin{figure}[#1]%
}{%
  \end{figure}%
}

\title{Event-driven eligibility propagation in large sparse networks: efficiency shaped by biological realism}

\author[1,2]{Agnes Korcsak-Gorzo \orcid{0000-0001-6496-4616}} % chktex 8
\author[3]{Jesús A. Espinoza Valverde \orcid{0009-0000-3728-8018}} % chktex 8
\author[4]{Jonas Stapmanns \orcid{0000-0002-5611-909X}} % chktex 8
\author[1,6,7]{Hans Ekkehard Plesser \orcid{0000-0001-7843-5993}} % chktex 8
\author[1]{David Dahmen \orcid{0000-0002-7664-916X}} % chktex 8
\author[3]{Matthias Bolten \orcid{0000-0002-8682-7652}} % chktex 8
\author[1,5,*]{Sacha J. van Albada \orcid{0000-0003-0682-4855}} % chktex 8
\author[1,2,8,9,*]{Markus Diesmann \orcid{0000-0002-2308-5727}} % chktex 8

\affil[1]{Institute for Advanced Simulation 6 (IAS-6), Jülich Research Centre, Jülich, Germany}
\affil[2]{Department of Physics, Faculty 1, RWTH Aachen University, Aachen, Germany}
\affil[3]{Department of Mathematics and Science, University of Wuppertal, Wuppertal, Germany}
\affil[4]{Department of Physiology, University of Bern, Bern, Switzerland}
\affil[5]{Institute of Zoology, University of Cologne, Cologne, Germany}
\affil[6]{Department of Data Science, Faculty of Science and Technology, Norwegian University of Life Sciences, Aas, Norway}
\affil[7]{Käte Hamburger Kolleg, RWTH Aachen University, Aachen, Germany}
\affil[8]{JARA-Institute Brain Structure-Function Relationships (INM-10), Jülich Research Centre, Jülich, Germany}
\affil[9]{Department of Psychiatry, Psychotherapy and Psychosomatics, Medical Faculty, RWTH Aachen University, Aachen, Germany}
\affil[*]{joint last}

\begin{document}

\maketitle

\section{Abstract}
  Despite remarkable technological advances, AI systems may still benefit from biological principles, such as recurrent connectivity and energy-efficient mechanisms. Drawing inspiration from the brain, we present a biologically plausible extension of the eligibility propagation (e-prop) learning rule for recurrent spiking networks. By translating the time-driven update scheme into an event-driven one, we integrate the learning rule into a simulation platform for large-scale spiking neural networks and demonstrate its applicability to tasks such as neuromorphic MNIST\@. We extend the model with prominent biological features such as continuous dynamics and weight updates, strict locality, and sparse connectivity. Our results show that biologically grounded constraints can inform the design of computationally efficient AI algorithms, offering scalability to millions of neurons without compromising learning performance. This work bridges machine learning and computational neuroscience, paving the way for sustainable, biologically inspired AI systems while advancing our understanding of brain-like learning.

\section{Introduction}
  Artificial intelligence is transforming many aspects of life, from large language models in chatbots to computer vision in security cameras and generative models for content creation. As neural networks and machine learning algorithms grow in scale and complexity, their energy demand increases, posing a challenge: optimizing them for energy efficiency while maintaining performance. A promising approach is to draw inspiration from the human brain, which operates with remarkable efficiency. Brain-inspired learning models not only deepen our understanding of neural processes but also offer opportunities for sustainable AI \supercite{roy2019towards}.

  One class of models supported by extensive evidence \supercite{gerstner2018eligibility,shouval2025eligibility} are three-factor models, which combine classical Hebbian learning with an additional modulatory signal \supercite{fremaux2016neuromodulated,magee2020synaptic}.
  The first and second factors --- the pre- and postsynaptic activities --- form the Hebbian component, determining whether a synapse is eligible to be modified by a third factor, for instance neuromodulation. An example is eligibility propagation (e-prop) \supercite{bellec2020solution}, a biologically plausible plasticity rule for recurrent spiking neural networks (SNNs)
  in which the Hebbian part is represented by an eligibility trace. E-prop is an online learning algorithm that approximates the exact gradients computed by backpropagation through time (BPTT) \supercite{werbos1990backpropagation}, achieving a good trade-off between performance and biological plausibility by avoiding nonlocality, time blocking, and the requirement for symmetric feedback weights. The concurrently developed Random-Feedback Online Learning (RFLO) algorithm\supercite{murray2019local}  shares the same core idea --- eligibility traces modulated by error broadcasts for online weight updates --- but operates on rate-based neurons. Here, we focus on supervised e-prop variants, alongside which reward-based variants also exist. The e-prop algorithm was originally implemented in TensorFlow \supercite{abadi2016tensorflow} with time-driven weight updates computed synchronously at each computational simulation time step, hereafter referred to as a ``step''.

  Time-driven weight update methods waste resources by ignoring the sparse nature of spiking activity. In biological networks, connectivity is sparse and neurons spike infrequently, with inter-spike intervals far exceeding the time scale of neuronal dynamics. Event-driven algorithms, where synaptic weight updates are computed asynchronously and only when an event occurs at the corresponding synapse, exploit this sparsity for computational advantage \supercite{morrison2007spike}. Event-driven updates reduce computation in sparse settings, improving e-prop scalability and, by leveraging advances in high-performance computing, enabling large-scale simulations. Such brain-scale and brain-inspired simulations are of interest to computational neuroscience for understanding learning and to machine learning for solving large-scale tasks.

  We present a computationally accurate, efficient, and scalable version of e-prop with event-driven weight updates and further biologically inspired features, and benchmark its performance. First, we present an event-driven algorithm for synaptic weight updates (\cref{sec:porting}). In \cref{sec:tasks}, we reproduce regression and classification proof-of-concept tasks from the original e-prop publication  \supercite{bellec2020solution}, showing that the event-driven scheme replicates time-driven results. Subsequently, we introduce an e-prop model with additional biological features (\cref{sec:biological-features}). We evaluate the performance on neuromorphic MNIST (N-MNIST) \supercite{orchard2015converting} and the scalability of both event-driven e-prop models in simulations. Finally, \cref{sec:methods} details the mathematical formulations of the time- and event-driven algorithms.

  Preliminary results were presented in abstract form \supercite{espinoza2024event_poster}.

\section{Results}

  \subsection{From time-driven to event-driven e-prop}\label{sec:porting}

    In artificial neural networks (ANNs), all components, including neurons and synapses, are updated synchronously at each step, making matrix multiplications well suited, as in TensorFlow \supercite{abadi2016tensorflow} and PyTorch \supercite{ansel2024pytorch}. In contrast, SNNs benefit from algorithms tailored to their temporal dynamics.

    \subsubsection{The case for event-driven synapse updates}

      Differential equations describing neuron dynamics are typically integrated with a step size of \SI{0.1}{\milli\second}, setting the smallest relevant timescale. Neurons communicate via spikes carrying information through timing alone. Spike propagation times, or synaptic transmission delays, vary and are often an order of magnitude larger than the simulation step outside local cortical networks. Firing rates depend, among other aspects, on brain state and neuron type and often follow a lognormal distribution, which peaks in human cortex around \SI{3}{\spikepersecond} \supercite{buzsaki2014log}. At such a rate, approximately \SI{3000}{} steps separate two spikes in simulations. Actual rates may be even lower due to underrepresentation of rarely active neurons in electrophysiological recordings\supercite{shoham2006silent}.

      As a neuron has roughly \SI{10000}{} incoming synapses, it processes an input spike about every \SI{0.03}{\milli\second}, close to the step required for its internal dynamics. Thus, incoming spikes are rare from a synapse's perspective but not from that of a neuron. Given the differing timescales and object counts, treating neurons and synapses as distinct entities is computationally advantageous. Accordingly, some SNN simulation algorithms adopt a hybrid event- and time-driven approach: spike communication and synaptic processing are event-driven, while neuron state updates are time-driven. Extensive research on such hybrid algorithms \supercite{morrison2005advancing,morrison2007spike,morrison2008phenomenological} demonstrates their suitability for parallel and distributed computing \supercite{jordan2018extremely,kurth2022sub}.

    \subsubsection{Event-driven weight updates in e-prop}
      The e-prop learning rule for SNNs relies on information locally available at the synapse. Based on their number, synapses form the primary computational bottleneck, followed by recurrent and output neurons. This makes the rule well suited to transition from the time-driven strategy to a hybrid approach that exploits the spatiotemporal sparsity of large-scale SNNs.

      Each synapse maintains an eligibility trace that captures how the filtered presynaptic spikes \(z_i^{t-1}\) and the postsynaptic surrogate gradient \(\psi_j^t\) of neuron \(j\) influence its contribution to learning. Global learning signals originating from output neurons represent the network's overall error. To update the weights, synapses multiply the eligibility trace by the summed learning signal \(L_j^t\) reaching neuron \(j\) (\cref{eq:gradient-rec,fig:technical_implementation-a}). The eligibility trace provides local credit assignment, while the global signal delivers feedback, enabling online, biologically plausible learning. Exploiting an established archiving framework \supercite{morrison2007spike,stapmanns2021event}, in our hybrid scheme event-driven synapses store \(z_i^{t-1}\) and time-driven neurons store \(L_j^t\) and \(\psi_j^t\) (see \cref{fig:technical_implementation-a,sec:architecture}).

      \begin{figure}[htbp]
        \centering
        \includegraphics[width=\textwidth]{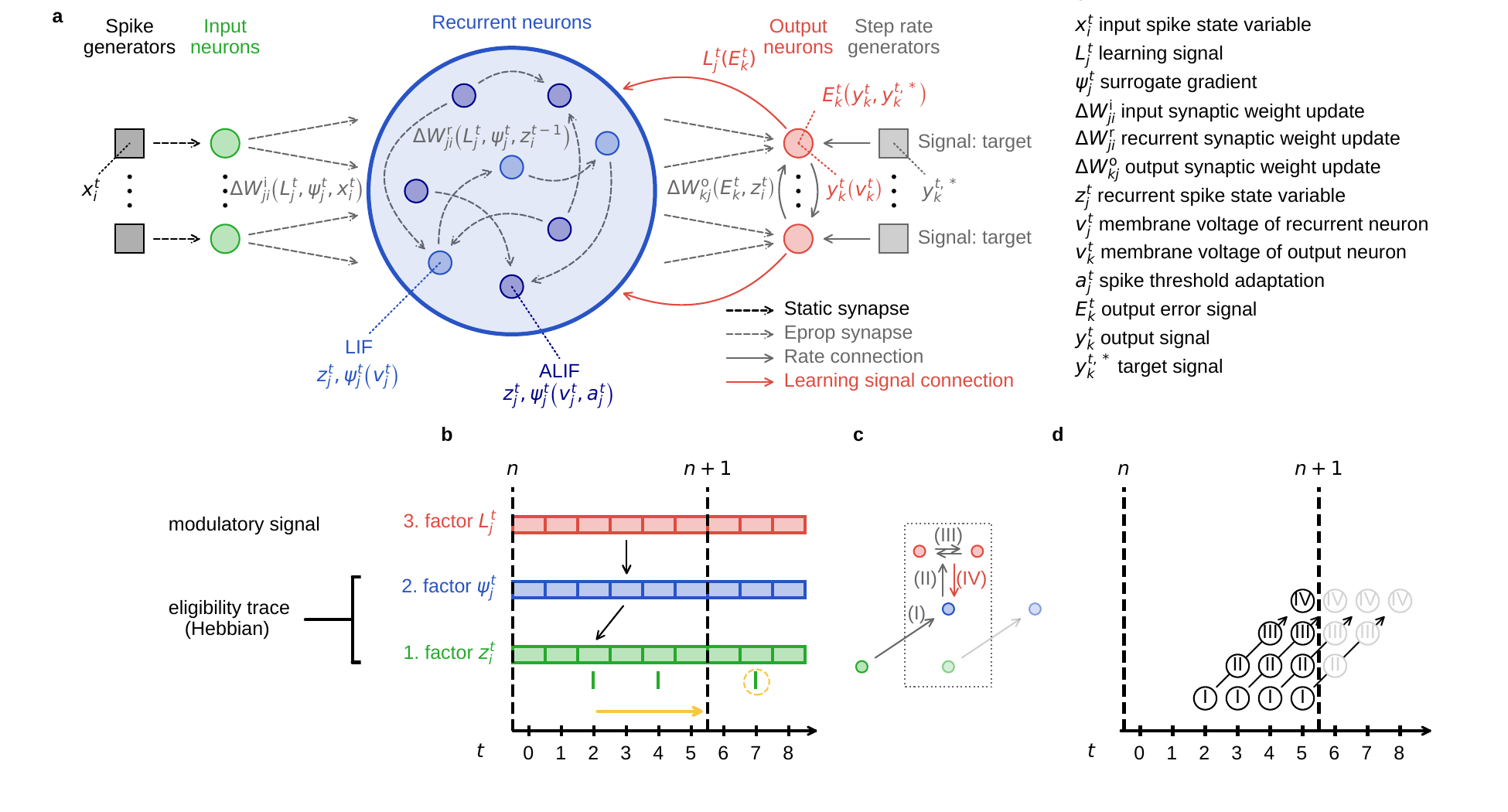}
        \phantomsubfigure{fig:technical_implementation-a}%
        \phantomsubfigure{fig:technical_implementation-b}%
        \phantomsubfigure{fig:technical_implementation-c}%
        \phantomsubfigure{fig:technical_implementation-d}%
        \caption[Mathematical basis and technical implementation of e-prop with event-driven weight updates]{\captiontitle{Mathematical basis and technical implementation of e-prop with event-driven weight updates.}
        \subfiglabel{a} Network layout and dynamic variables for weight updates in input, recurrent, and output synapses. The recurrent network may include LIF, ALIF, or both neuron types.
        \subfiglabel{b} The first spike (highlighted with yellow circle) of the neural response to input data sample \(n+1\) triggers the retrieval of the archived history for the previous sample \(n\) (yellow arrow) and the computation of the corresponding gradients. Arrows indicate the surrogate gradient entry \(\psi_j^t\) and presynaptic spike \(z_i^{t-1}\) associated with a learning signal entry \(L_j^t\).
        \subfiglabel{c} Propagation of spikes and signals within a single step (dotted box). Over e-prop synapses (gray arrows): (I) Transmit spikes from input neurons (green circles) to recurrent neurons (blue circles), (II) transmit spikes to output neurons (red circles). (III) Transmit signals (gray arrows) between output neurons to compute the softmax. (IV) Send the learning signal (red arrow) from the output layer to the recurrent layer.
        \subfiglabel{d} Representation of transmission I-IV as a pipeline across four steps. The number of incomplete operations (gray circles) at the boundary (dashed black line) increases with pipeline depth. In this case, three learning signals have not arrived yet at update time, corresponding to the pipeline depth minus one.
        }\label{fig:technical_implementation}
      \end{figure}

      During training, different data samples are presented to the network as input spike patterns of duration \(T\) (typically \SIrange{1}{2}{\second}), which are processed in mini-batches of size \(N\). The time-driven implementation accumulates gradients at every step, updates the weights using the averaged gradients after each mini-batch of duration \(NT\), referred to as an iteration. In contrast, the event-driven implementation computes and accumulates the gradient at the first spike after each sample (\cref{fig:technical_implementation-b}), averages the accumulated gradients and updates the weights at the first spike after the iteration, ensuring the first spike requiring the updated weight uses the correct value. Thus, the time-driven model updates all synapses synchronously, whereas the event-driven model clusters updates at the start of each iteration. In sparse spiking scenarios, the first spike after a completed sample may occur one or more samples later; in such cases, the synapse retrieves postsynaptic histories to compute the update for the sample in which the previous spike occurred. In samples without incoming spikes, the presynaptic factor remains zero, producing no update, and the algorithm disregards those histories.

      The loss \(\loss \) of one sample is the sum of incremental losses \(l^t\) at each step \(t\), computed from the history values \(z_i^{t-1}\), \(\psi_j^t\), and \(L_j^t\) (see \cref{sec:gradients,eq:accumulated-gradients}). Simple gradient descent or the Adam optimizer \supercite{kingma2017adam} (\cref{eq:adam}) compute the weight update from the average of gradients accumulated over a mini-batch. To free memory, the algorithm clears histories of all samples where all synapses have retrieved their data or had no presynaptic spikes (\cref{sec:history-management}).

    \subsubsection{Transmission delays}\label{sec:transmission-delays}

      For simplicity, the original derivation of e-prop \supercite{bellec2020solution} considers only recurrent delays, fixed to the solver's step. In their supplement, the authors generalize the formulas to non-uniform and arbitrary input and recurrent delays. However, transmissions to, within, and from the output layer are still treated as instantaneous on the timescale of the step (see \cref{fig:technical_implementation-c}) as reflected in the shared time index \( t \) of the respective variable pairs: recurrent state vector \(\vb*{z}^t\) and output signal vector \(\vb*{y}^t\) in \cref{eq:leaky_integrator}, numerator and denominator of the softmax function in \cref{eq:softmax}, and \(L_j^t\) and \(\psi_j^t\) in \cref{eq:filtered-eligibility-trace}. The latter shared time index introduces a misalignment, as \(L_j^t\), computed in the output layer --- remote from the recurrent neurons --- does not yet carry the effect of recurrent spikes and the associated membrane voltages underlying \(\psi_j^t\), which are first reflected by \(L_j^{t+1}\).

      Zero-delay transmissions conflict with empirical evidence of delays in neurobiological processes \supercite{sabatini1999timing} and their critical role in brain information processing and representation \supercite{jirsa2004connectivity}. They also challenge neural simulators, which usually require at least one step for spike and signal transmission. To address this, our implementation compensates for delays by synchronizing histories of the factors involved in weight updates (see black arrows connecting history entries in \cref{fig:technical_implementation-b}). We also decouple the transmission delay from the step for conceptual clarity, enabling shorter steps.

      In frameworks with non-zero transmission delay, instantaneous transmissions can be interpreted as a pipeline unrolled over multiple steps\supercite{allan1995software}. The pipeline of instantaneous transmissions can be unrolled over multiple steps when implemented in a framework with non-zero transmission delays \supercite{allan1995software}. Under a one-step transmission delay, the three originally instantaneous transmissions result in three missing learning signal values at the end of a sample (\cref{fig:technical_implementation-d}). We recover the third by accumulating gradients over a left-open interval and address the remaining signals and delay handling in later sections.

  \subsection{Supervised benchmark tasks with event-driven e-prop}\label{sec:tasks}

    In the following, we use three established tasks to verify our event-driven algorithm with respect to accuracy and to evaluate further algorithmic changes.

    \subsubsection{Pattern generation}\label{sec:proof-of-concept-regression}

      As a first proof of concept, we reproduce the pattern generation regression task \supercite{bellec2020solution} with the event-driven strategy (for background see \cref{sec:pattern-generation} and for dynamics \cref{fig:proof_of_concept_regression-a}). The task uses mean squared error (\cref{eq:mean-squared-error}) optimized with gradient descent (\cref{eq:gradient-descent}). Following the original implementation, we train all weights, but training output weights suffices (\cref{fig:proof_of_concept_regression-b}). The loss time course of the event-driven implementation matches that of the time-driven implementation (\cref{fig:accuracy_comparison_event_time_driven-a,fig:accuracy_comparison_event_time_driven-b}). Inspired by the literature \supercite{laje2013robust}, we also train two-dimensional patterns (\cref{fig:additional-tasks-a,fig:additional-tasks-b}) using two output neurons to encode the horizontal and vertical coordinates of a signal.

      \begin{figure}[htbp]
        \centering
        \includegraphics[width=\textwidth]{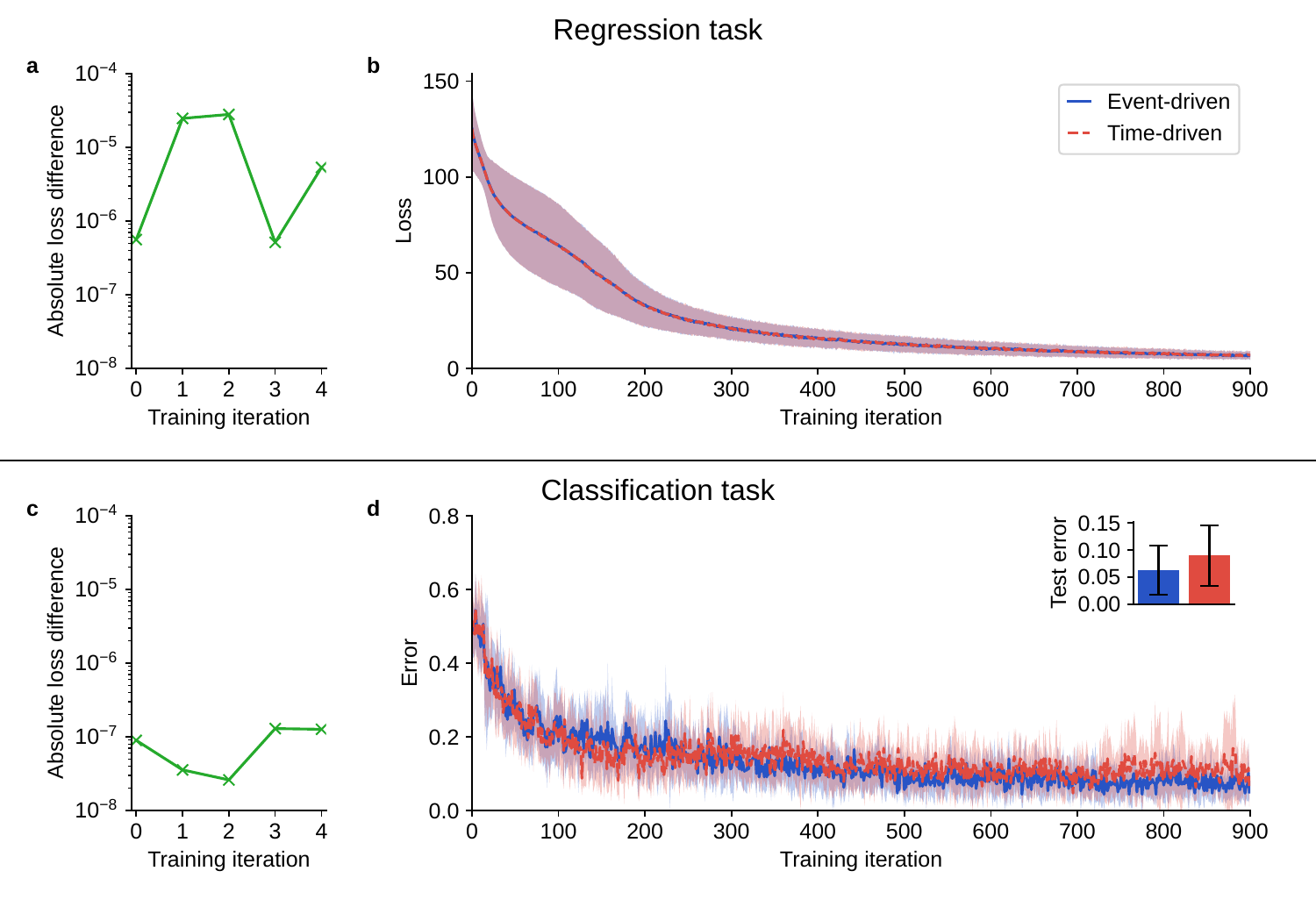}
        \phantomsubfigure{fig:accuracy_comparison_event_time_driven-a}%
        \phantomsubfigure{fig:accuracy_comparison_event_time_driven-b}%
        \phantomsubfigure{fig:accuracy_comparison_event_time_driven-c}%
        \phantomsubfigure{fig:accuracy_comparison_event_time_driven-d}%
        \caption[Comparison of learning performance between time-driven and event-driven models.]{\captiontitle{Comparison of learning performance between time-driven and event-driven models.}
          Regression task (pattern generation): \subfiglabel{a} Absolute difference in loss between the two models during the first \SI{4}{} training iterations with a mini-batch size of \SI{1}{} in a single trial and \subfiglabel{b} loss of both models over \SI{900}{} training iterations with a mini-batch size of \SI{1}{} averaged over \SI{10}{} trials. Classification task (evidence accumulation): \subfiglabel{c} Absolute difference in loss between the two models during the first \SI{4}{} training iterations with a mini-batch size of \SI{1}{} in a single trial and \subfiglabel{d} prediction error of both models over \SI{900}{} training iterations with a mini-batch size of \SI{32}{} averaged over \SI{10}{} trials. Curves represent the mean and shaded areas the standard deviation across \SI{10}{} trials with different random seeds. Bars in the inset show the mean across all trials and \SI{10}{} test iterations per trial, and error bars represent the combined standard deviation, calculated as the square root of the sum of within-trial and between-trial variances.
        }\label{fig:accuracy_comparison_event_time_driven}
      \end{figure}

    \subsubsection{Evidence accumulation}\label{sec:proof-of-concept-classification}

      As a second proof of concept, we reproduce the evidence accumulation classification task \supercite{bellec2020solution} (for background see \cref{sec:evidence-accumulation}, for dynamics \cref{fig:proof_of_concept_classification-a}, for weight distributions \cref{fig:proof_of_concept_classification-b}), with weight updates optimized by the Adam algorithm (\cref{eq:adam}). Here, mini-batch learning improves performance: in the time-driven model it is implemented by running multiple network copies, equal to the mini-batch size, in parallel with identical initialization but different task examples; after each iteration, the weight update based on the averaged gradients is applied across all networks.

      For biologically realistic mini-batch learning, we process samples sequentially and apply the weight update after the completion of the mini-batch using the averaged gradients. Parallel and sequential processing are equivalent when neuron and e-prop trace dynamics are fully reset at the end of each mini-batch. In \cref{sec:continuous-dynamics}, we discuss removing the biologically unrealistic reset mechanism.

      To solve a classification task, each output neuron represents one class; here, two neurons correspond to the two options. The cross-entropy loss (\cref{eq:cross-entropy-loss}) is computed with a softmax function (\cref{eq:softmax}), which converts output neuron membrane voltages into probabilities by dividing the exponential of each voltage by the sum of all exponentials. This requires continuous access to all output voltages. We enable this by adding rate connections between output neurons, introducing an extra step. Evidence indicates such divisive normalization mechanisms exist in the brain \supercite{carandini1994summation,wilson2012division}.

      To directly compare the event-driven and time-driven models, we record the loss from a single trial to measure how much each output neuron's membrane voltage deviates from its target value (\cref{fig:accuracy_comparison_event_time_driven-c}). With a mini-batch size of \SI{1}{}, we evaluate the network after each sample. For the first few samples, differences remain near numerical precision for floating-point arithmetic, showing that the event-driven model closely matches the time-driven loss in the classification task.

      To compare the classification performance of the two models, we use the prediction error as the standard metric, measuring how often the network correctly classifies samples within each mini-batch. With a mini-batch size of \SI{32}{}, the prediction error decreases at a similar rate for the time-driven and event-driven models during the initial training iterations. Larger deviations emerge later, but they do not compromise overall learning success (\cref{fig:accuracy_comparison_event_time_driven-d}). These deviations arise from minor numerical discrepancies in floating-point arithmetic, which cause small differences in the computed gradients between the event-driven and time-driven models. The resulting variations in updated weights can shift a neuron's membrane voltage enough for it to cross the spike threshold in one simulation but not in the other.

      To analyze this effect, we record the spike pattern from one simulation and compare it to a second in which we introduce a single additional spike (\cref{fig:spike-perturbation-a}). Because the recurrent network operates near the edge of chaos, such a perturbation can trigger a cascade of spike differences. Consequently, the spike patterns of the perturbed and unperturbed simulations diverge, amplifying loss deviations that remain small overall (\cref{fig:spike-perturbation-b}) but are still measurable (\cref{fig:spike-perturbation-c}).

    \subsubsection{Neuromorphic MNIST}\label{sec:nmnist}
\begin{figure}[htbp]
        \centering
        \includegraphics[width=0.6\textwidth]{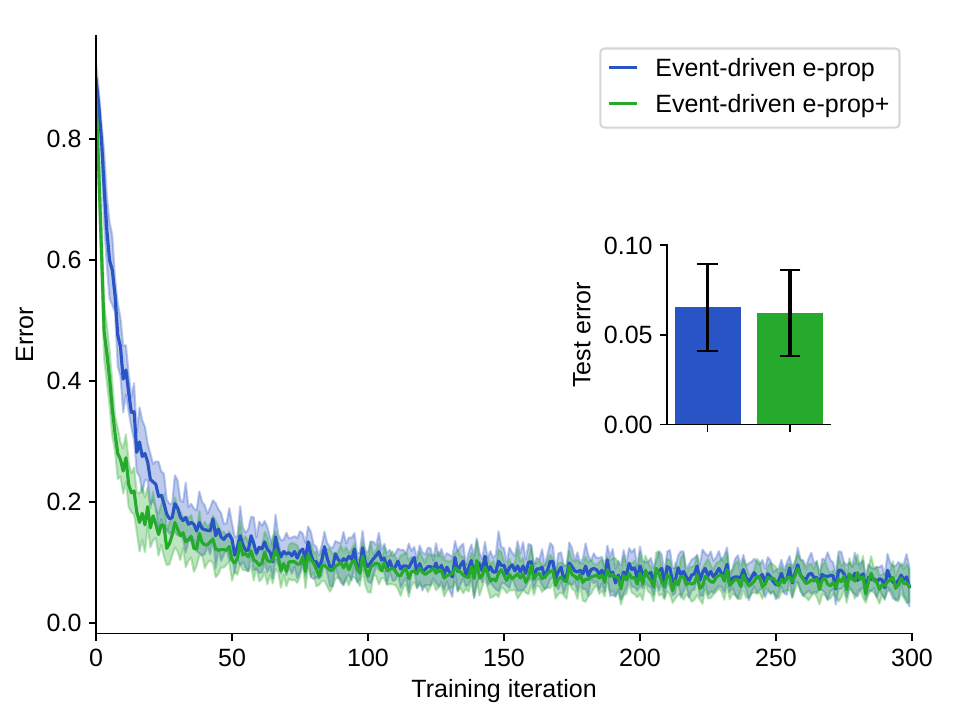}
        \caption[Learning performance of event-driven e-prop models]{\captiontitle{Learning performance of event-driven e-prop models.}
          Learning performance measured as prediction errors on the N-MNIST task using event-driven e-prop models. Curves represent the mean and shaded areas the standard deviation across \SI{10}{} trials with different random seeds. Bars in the inset show the mean across all trials and \SI{10}{} test iterations per trial, and error bars represent the combined standard deviation, calculated as the square root of the sum of within-trial and between-trial variances. The convergence speed and test error of \epropplus{} match those of e-prop, confirming that learning remains effective. Learning converges slightly faster for \epropplus{}, possibly because the identical parameters across both models produce dynamics more favorable to learning. A precise conclusion would require fair parameter tuning for both models.
        }\label{fig:accuracy_comparison_with_without_bio}
      \end{figure}

      As a more challenging benchmark, we implement a classification task on the basis of the N-MNIST dataset \supercite{orchard2015converting} (\cref{sec:neuromorphic-mnist}). The solver measures learning performance by the classification error using cross-entropy loss, and optimizes performance with gradient descent and a mini-batch size of \SI{1}{}. The time course of the prediction error in \cref{fig:accuracy_comparison_with_without_bio} demonstrates successful learning of the N-MNIST task with event-driven e-prop. The supplement includes the time courses of the dynamic variables (\cref{fig:nmnist-original-traces-a}) and the weight distributions (\cref{fig:nmnist-original-traces-b}) before and after training. We use this task in the following experiments in \cref{sec:biological-features} to evaluate modifications of event-driven e-prop that increase the compatibility of the algorithm with biological constraints.

  \subsection{Event-driven e-prop with additional biological features}\label{sec:biological-features}
\begin{figure}[htbp]
      \centering
      \includegraphics[width=\textwidth]{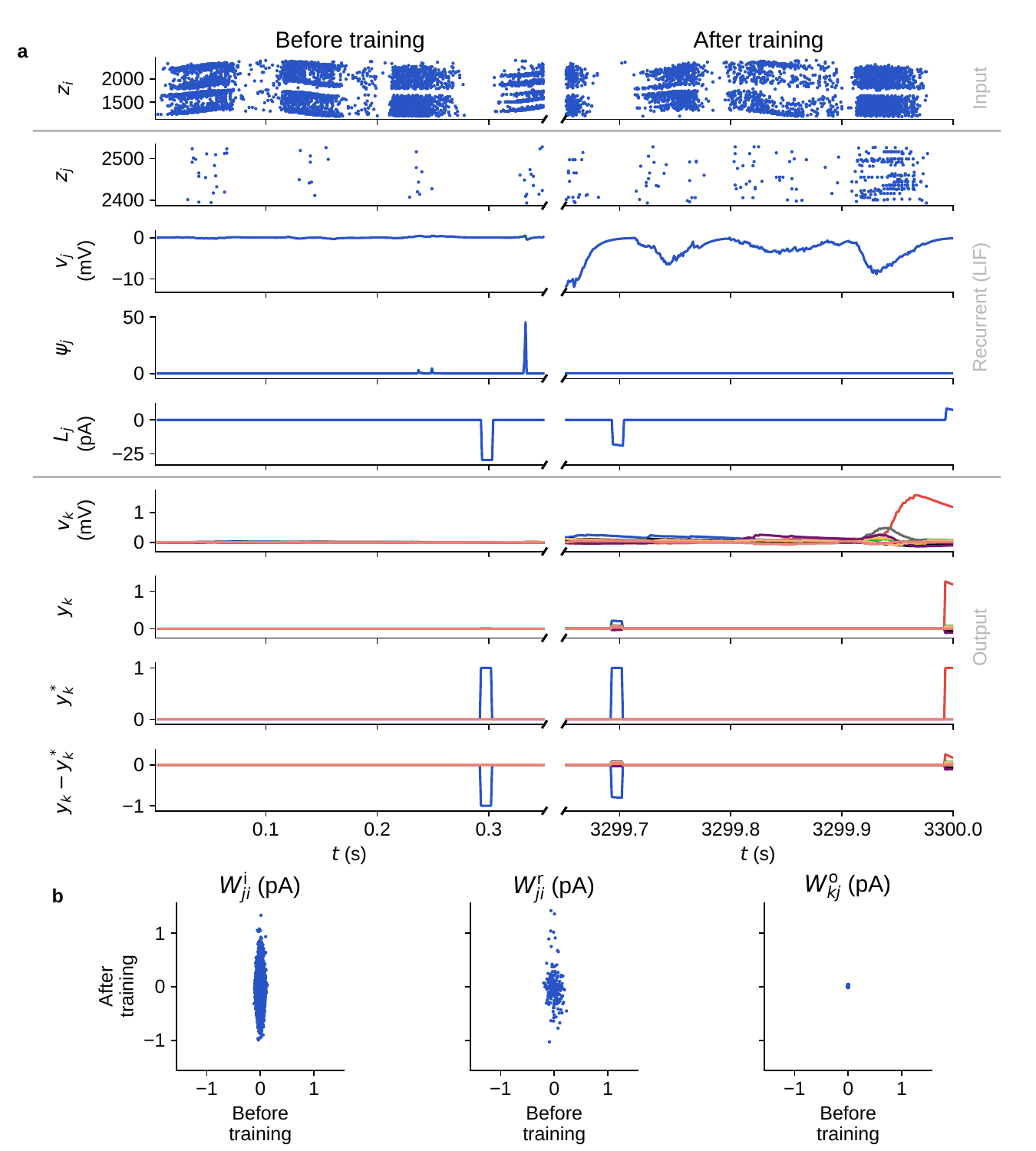}
      \phantomsubfigure{fig:nmnist-bio-traces-a}%
      \phantomsubfigure{fig:nmnist-bio-traces-b}%
      \caption[Dynamics and weight distribution before and after training N-MNIST using \epropplus]{\captiontitle{Dynamics and weight distribution before and after training N-MNIST using \epropplus.}
      \subfiglabel{a} Time traces of dynamic variables before and after training. Spike states \(z_i\) for the input neurons; spike state \(z_j\), membrane voltage \(v_j\), surrogate gradient \(\psi_j\), and learning signal \(L_j\) for an example recurrent LIF neuron; and membrane voltages \(v_k\), target signals \(y_k^{*}\), readout signals \(y_k\), and their differences \(y_k - y_k^*\) for the ten output neurons. After training, the output neuron with the highest membrane voltage (and thus the highest readout signal) matches the target output neuron (red curves), correctly predicting the class.
      \subfiglabel{b} Distributions of input, recurrent, and output weights. Points on the diagonal indicate no change. The largest changes occur in the input and recurrent weights.
      }\label{fig:nmnist-bio-traces}
    \end{figure}

    Several components of e-prop are adapted from machine learning methods and favor mathematical simplicity and rigor. In this section, we shift the trade-off in these components toward greater biological realism and refer to the resulting event-driven e-prop model as \epropplus{} for brevity.\ \cref{fig:nmnist-bio-traces} compares network activity and configuration before and after training. In \cref{fig:nmnist-bio-traces-a}, \epropplus{} dynamics show that after training, the output neuron whose readout signal attains the largest values aligns with the target, correctly predicting the class. Weight distributions in \cref{fig:nmnist-bio-traces-b} show that this outcome arises mainly from adjustments to input and recurrent weights.

\begin{figure}[htbp]
      \centering
      \includegraphics[width=\textwidth]{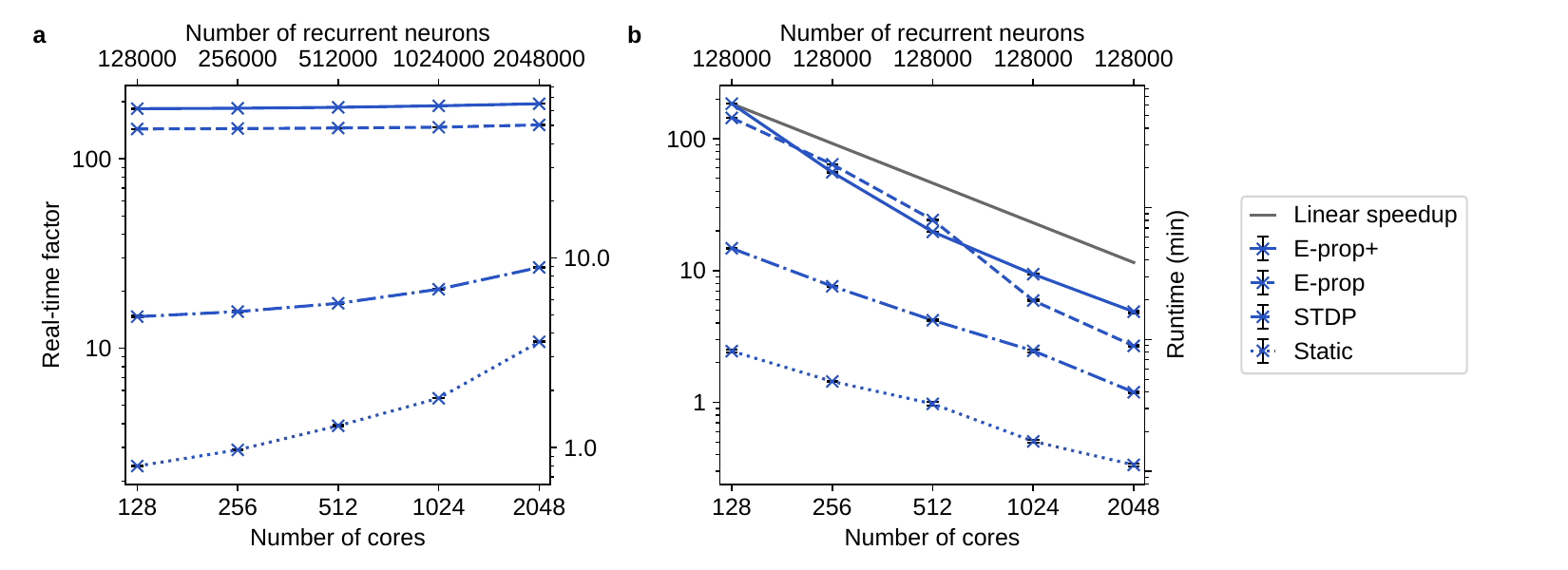}
      \phantomsubfigure{fig:scaling-a}%
      \phantomsubfigure{fig:scaling-b}%
      \caption[Scaling of e-prop models]{\captiontitle{Scaling of e-prop models.}
        \subfiglabel{a} Weak scaling and \subfiglabel{b} strong scaling results shown as wall-clock runtime of the simulation excluding the network-building phase, and the real-time factor defined as the ratio of simulated biological time to runtime, both as a function of the number of cores. The results reveal sub-linear strong scaling, with runtime differences diminishing as the number of cores increases, and near-linear weak scaling. The gray line indicates ideal linear speedup where runtime decreases inversely with number of cores. As expected, plasticity calculations increase runtime compared to static synapses, and the third factor in e-prop adds overhead relative to the two-factor STDP rule. Points and error bars show mean and standard deviation across runs, though the error bars are barely visible due to low variance.
      }\label{fig:scaling}
    \end{figure}

    \cref{fig:accuracy_comparison_with_without_bio} shows that \epropplus{} maintains learning effectiveness, matching e-prop in convergence speed and final test error. Runtime measurements under varying degrees of parallelization and network sizes --- up to \SI{2}{} million neurons --- demonstrate good scalability of both event-driven e-prop models, with super-linear strong scaling and near-ideal weak scaling (see \cref{fig:scaling,sec:scaling-methods}).

    \subsubsection{Dynamic firing rate regularization}\label{sec:dynamic-firing-rate-regularization}

      In the time-driven \supercite{bellec2020solution} and our event-driven model, standard firing rate regularization depends on the sample duration (\cref{eq:firing-rate-regularization}). A dynamic variant in the time-driven model removes this dependence by using a cumulative average of the spike count since the start of the iteration normalized by the current step (\cref{eq:sma-loss,eq:sma-rate}). We replace the standard cumulative average with an exponential moving average by making the dynamic filter variable constant (\cref{eq:ema-rate}). This removes explicit dependence on time and a fixed origin and, unlike the standard average that weights all spikes equally, emphasizes recent spikes, making it more responsive to dynamic activity changes.

    \subsubsection{Classification via mean squared error}\label{sec:classification-via-mse}

      For classification tasks, the original implementation uses cross-entropy loss (\cref{eq:cross-entropy-loss}), requiring a softmax output (\cref{eq:softmax}), which introduces an extra step for communication between output neurons to calculate the denominator. Since squared error loss performs comparably to cross-entropy across various neural network architectures and benchmarks \supercite{hui2021evaluation}, we instead employ a temporal mean squared error (\cref{eq:temporal-mse}) which avoids the extra communication. As discussed in \cref{sec:transmission-delays,fig:technical_implementation-d}, introducing a nonzero delay to previously instantaneous transmissions results in three learning signal values yet to arrive at update time. Using the temporal mean squared error reduces this number by one.

    \subsubsection{Continuous dynamics}\label{sec:continuous-dynamics}

      The original algorithm \supercite{bellec2020solution} resets the dynamic variables of neurons and the filtered e-prop traces to zero after each weight update to prevent residual activity from one iteration affecting weight updates of the next. In experiments without resets, we observe that they are not critical for maintaining learning performance, suggesting minimal impact from sample interference.

    \subsubsection{Learning window signal generator}\label{sec:learning-window-signal-generator}

      The evidence accumulation task requires a learning window that opens during the final few hundred milliseconds of each iteration (\cref{sec:evidence-accumulation}). To enable flexible, iteration-independent control, we introduce a learning-window signal generator that sends a signal of value \SI{1}{} to the connected output neurons when the learning window is open, and \SI{0}{} when it is closed. The output neurons multiply the learning signal by this value, thereby enabling or disabling plasticity accordingly. In our reference implementation, this is realized similarly to the target signal generators (\cref{fig:technical_implementation-a}), using rate generators connected via rate connections.

    \subsubsection{Weight updates with every spike}\label{sec:weight-updates-with-every-spike}

      Computing weight updates over a fixed time interval imposes two challenging requirements: all training samples must have the same duration, and a central clock must broadcast update times to all synapses and neurons. To avoid the biological implausibility of relying on a fixed temporal grid for these computations, we introduce a scheme inspired by truncated algorithms (\cref{sec:locality-causality-onlineness}). In this scheme, each spike triggers a weight update based on the time since the previous spike (see \cref{sec:algorithms}). This aligns with plasticity mechanisms such as short-term plasticity \supercite{zucker2002short,tsodyks2000synchrony} and spike-timing dependent plasticity (STDP) \supercite{bi1998synaptic,morrison2007spike}, a synaptic learning rule where the strength of a connection depends on the precise timing of pre- and postsynaptic spikes. For two consecutive spikes in a synapse, the algorithm computes the weight update up to the most recent learning signal, based on the e-prop history within the left-open inter-spike interval. Unlike time-driven e-prop, where weights remain fixed throughout a mini-batch, and event-driven e-prop, where they remain fixed between the first spikes of successive mini-batches, this scheme updates the weights with every transmitted spike.

      Together, the concepts of \cref{sec:dynamic-firing-rate-regularization,sec:classification-via-mse,sec:continuous-dynamics,sec:learning-window-signal-generator} along with delays between recurrent and output layer (\cref{sec:generalized-delays}), support the elimination of clocked weight computations. As a result, the new scheme enables learning from samples of varying duration.

    \subsubsection{Eligibility trace filter decoupled from output time constant}\label{sec:weight-update-agnostic-of-output-time-constant}

      In the derivation of the update rule for synaptic weights in the time-driven model \supercite{bellec2020solution}, the time constant of the output neuron (\cref{eq:leaky_integrator}) enters the filter characteristics of the eligibility trace (\cref{eq:filtered-eligibility-trace}). Consequently, for synapses to compute their weight updates, they must know the time constant of the output neuron, which violates the principle of locality. In our experiment, removing the filter entirely in a regression task improves learning performance (see \cref{fig:etrace-filter}). This suggests that the filter of the eligibility trace can be set independently, so that synapses no longer need to know the time constant of the output neuron.

    \subsubsection{Smooth surrogate gradient function}\label{sec:smooth-surrogate-gradient-function}

      The time-driven model employs a piecewise linear surrogate gradient (\cref{eq:piecewise-linear-surrogate-gradient}). Many alternative functions exist in the literature\supercite{neftci2019surrogate}, which, when adjusted for height and width, yield similar shapes (\cref{fig:surrogate-gradient-functions-a}) and comparable performance, further improved by tuning these parameters (see \cref{fig:surrogate-gradient-functions-b}). These findings agree with studies showing that learning is robust against shape variations of the surrogate gradient \supercite{zenke2021remarkable}. Thus, replacing the piecewise linear function by a smoother exponential surrogate gradient \supercite{shrestha2018slayer} enhances biological realism and mathematical simplicity without compromising performance (\cref{eq:exponential-surrogate-gradient}).

      %\subsubsection{Weight constraints and sparse connectivity}\label{sec:sparse-connectivity-and-weight-constraints}

      %  Our experiments show that biologically realistic sparse recurrent connections (\cref{sec:bio-conn-weights}) reduce computational load without compromising learning performance. We further investigate biological weight constraints using \epropplus{}, enforcing fixed weight signs separately for input, recurrent, and output weights, as well as for all combined, and implementing Dale's law with an excitatory-to-inhibitory ratio of 4:1.

\section{Discussion}
  In this study, we extend e-prop, a biologically plausible three-factor learning rule for SNNs. As a reference, we embed the new algorithm in the open-source SNN simulation code NEST, optimized for distributed large-scale network simulations. All quantitative data reported here were obtained with this implementation. The code (\url{https://github.com/nest/nest-simulator/pull/2867}) is partly available in NEST release 3.7 \supercite{espinoza2024nest}, with additional functionality (\url{https://github.com/nest/nest-simulator/pull/3207}) in release 3.9 \supercite{terhorst2025nest}. To support accessibility, we provide detailed tutorials (\url{https://nest-simulator.readthedocs.io/en/stable/auto_examples/eprop_plasticity/index.html}). The conceptual and algorithmic work builds on previous efforts to implement three-factor rules in NEST \supercite{potjans2010enabling,stapmanns2021event} and is part of a long-term collaborative project aimed at advancing neural systems simulation technology \supercite{gewaltig2007nest}.

  We first adapt e-prop's original synchronous, time-driven weight updates to an asynchronous, event-driven framework and reproduce two supervised tasks from the original publication \supercite{bellec2020solution}. Minor numerical differences can trigger an extra spike in one implementation, causing cascades of downstream spikes and cumulative loss deviations, but overall learning success remains unaffected. Our model generalizes to additional tasks, including the N-MNIST benchmark, and can extend to others. Porting the algorithm from TensorFlow to NEST and adapting it to NEST's biologically grounded constraints yields insights such as the necessity of incorporating previously absent transmission delays and strict adherence to locality. The incorporation of transmission delays in hybrid simulation schemes with time-driven neuronal updates and event-driven synaptic communication furthermore allows the two corresponding time scales to be decoupled\supercite{morrison2005advancing}. Following the original e-prop implementation, our reference implementation uses the same value for the synaptic communication delay and the neuronal update step, resulting in information being sent at every step. Straightforward extensions that enable transmission on coarser time grids by buffering information \supercite{hahne2017integration} promise further efficiency gains, especially for larger delays or smaller neuronal update steps.

  Our e-prop implementation provides a foundation for implementing reward-based e-prop \supercite{bellec2020solution} and other three-factor learning rules in generic SNN simulations. Potential candidates include algorithms similar to Real-Time Recurrent Learning (RTRL) that are practical for neuromorphic hardware \supercite{zenke2021brain}. One example, EventProp \supercite{wunderlich2021event}, has recently been implemented in an event-driven manner on the neuromorphic hardware systems BrainScaleS \supercite{pehle2023event,billaudelle2020versatile} and SpiNNaker~2 \supercite{bena2024event,gonzalez2024spinnaker2}. Porting this class \supercite{marschall2020unified,tallec2017unbiased,roth2018kernel,mujika2018approximating,murray2019local,benzing2019optimal,jaderberg2017decoupled,lee2016training,menick2020practical,kaiser2020synaptic,bohnstingl2022online,lee2022exact,quintana2024etlp,wei2024event,klos2025smooth,liu2021cell} of online training algorithms for biologically plausible recurrent neural networks (\cref{sec:online-training-recurrent-nets}) to NEST may be best achieved by formalizing plasticity rules in NESTML\@\supercite{plotnikov2016nestml}, a domain-specific language for neuron and synapse models with automatic compilation to C++. Building on the port of neuromodulated STDP\supercite{potjans2010enabling,linssen2025nestml} --- a reward-based three-factor algorithm --- this approach enables greater flexibility for custom plasticity models.

  Future work could extend validation to broader tasks, network architectures, neuron parameters, and learning hyperparameters. Fair comparisons across frameworks require full optimization, a promising direction for further study that may be guided by recent work on parallelized gradient computation, which reduces runtime \supercite{baronig2025scalable}. The inherent energy efficiency of spike-based computation offers substantial savings and, combined with further biologically inspired mechanisms, may drive advances in machine learning technologies.

  This work contributes to efforts to port e-prop to various frameworks and substrates. E-prop has been implemented in mlGeNN \supercite{knight2022efficient,knight2023easy}, a spike-based machine learning library optimized for GPU-based sparse data structures via the pyGeNN simulator \supercite{knight2021pygenn}, demonstrating functionality on CPUs and GPUs\@. It has also been adapted for neuromorphic hardware, including SpiNNaker~1 \supercite{perrett2022online,rhodes2020real}, SpiNNaker~2 \supercite{rostami2022eprop,gonzalez2024spinnaker2}, and ReckOn \supercite{frenkel2022reckon}, with the SpiNNaker ports notably incorporating event-driven weight updates \supercite{perrett2022online,rostami2022eprop}. Comparing time and energy demands of neuromorphic and conventional solutions on the same task at fixed accuracy, similar to recent efforts \supercite{senk2025constructive}, could now be informative.

  In contrast to our study, which focuses on faithfully reproducing and extending the original e-prop implementation, these hardware implementation efforts primarily emphasize implementing the core mechanism and, in some cases, simplifying the algorithm to meet hardware constraints \supercite{frenkel2022reckon}. None demonstrates an exact reproduction of the original results, and only one explicitly compares learning performance \supercite{rostami2022eprop}. Tasks vary across studies: only one reproduces pattern generation \supercite{perrett2022online}, two address evidence accumulation \supercite{perrett2022online,frenkel2022reckon}, none use N-MNIST, though one explores sequential MNIST \supercite{knight2022efficient}. Further tasks include Google Speech Commands \supercite{rostami2022eprop}, DVS gestures \supercite{knight2022efficient}, spiking Heidelberg digits \supercite{knight2023easy,frenkel2022reckon}, and a synthetic behavioral dataset \supercite{frenkel2022reckon}. Most studies use networks of several hundred to a few thousand neurons, with only one explicitly analyzing scaling in processing time (measured in clock cycles) versus network size \supercite{rostami2022eprop}.

  Already the basic computational unit of the mammalian cerebral cortex, the so-called cortical microcircuit\supercite{douglas1989canonical,braitenberg1998cortex,potjans2014cell}
  contains two orders of magnitude more neurons than present-day network models using e-prop. Exploiting networks of natural size requires scalable algorithms. We present the first investigation of e-prop under weak and strong scaling, showing that our event-driven implementation scales to millions of neurons while capturing mammalian spatio-temporal sparsity. This broadens the range of potential applications and marks a milestone toward simulating plastic brain-scale networks. Our study emphasizes biological plausibility and examines how biological constraints affect efficiency and performance, whereas other works mainly optimize functional performance, energy use, and memory, often at the expense of biological plausibility.

  Integrated into a widely used simulation code for large-scale neuronal networks, this implementation provides a tool to study neuroscientifically relevant tasks, simulate behavior, and test learning hypotheses. It offers the potential for insights into how plasticity supports behavior and plays into neurological disorders while advancing machine learning for real-world applications.

\section{Methods}\label{sec:methods}

  In this section, we present the mathematical formulation of the original time-driven e-prop algorithm as introduced in Bellec et al. (2020) \supercite{bellec2020solution}, along with details of the network architecture, neuron models, tasks, and optimization schemes used to reproduce their experiments in our implementation as a proof of concept. We then describe our extensions. Occasionally, we make minor adjustments to the original mathematical notation to ensure consistency with our framework and for clarity, among others:
  \begin{notation}\label{not:exponent-notation}
    \renewcommand{\thenotation}{}%
    When \(t\) or related expressions serve as exponents, they are enclosed in parentheses, e.g.~\(x^{(t)}\), whereas a plain superscript, e.g.~\(x^t\), denotes an index.
  \end{notation}

  \subsection{Network architecture}

    The networks are recurrent SNNs, which consist of an input layer, a hidden layer with recurrent connections, and an output layer. The primary function of the hidden layer is to process input sequences over time, extract temporal patterns within the data, and generate corresponding spike sequences. Recurrence plays a crucial role by providing the network with memory, enabling it to retain information from previous steps. This allows the network to operate across multiple timescales, facilitating the effective modeling and processing of sequential or temporal data.

  \subsection{Neuron models}

    The \( J \) neurons in the hidden layer are modeled as leaky integrate-and-fire (LIF) neurons, in some cases with additional adaptation. The dynamics of these neurons receiving spikes from \( I \) input neurons are described by the update equation
    \begin{align}
      v_j^t & = \alpha v_j^{t-1} + \sum_{\substack{i=1 \\ i \neq j}}^J W_{ji}^\text{r} z_i^{t-1} + \sum_{i=1}^I W_{ji}^\text{i} x_i^t - z_j^{t-1} v_j^{\text{th},t},
      \label{eq:lif_update}
    \end{align}
    where \(j\) and \(i\) are neuron indices, \(W_{ji}^\text{r}\) and \(W_{ji}^\text{i}\) are the recurrent and input synaptic strengths, and \(\alpha = \exp\qty(-\frac{\Delta t}{\tau^\text{m}})\) represents the decay factor of the membrane voltage over time, with \(\tau^\text{m}\) denoting the membrane time constant. The binary spike state variable is given by
    \begin{align}
      z_j^t = H\qty(v_j^t - v_j^{\text{th},t}),
      \label{eq:binary-spike-state-variable}
    \end{align}
    and is 1 if the neuron spikes at time \(t\), and 0 otherwise. Note that \cref{eq:lif_update}, representing the time-driven approach\supercite{bellec2020solution}, assumes an instantaneous effect of the inputs \(x_i^t\) on the membrane potential \(v_j^t\). In our event-driven setup, the delay between input and recurrent neurons effectively yields \(x_i^{t-1}\). This is equivalent, however, since the dynamics commence with the first spike in the recurrent network, resulting only in a one-step forward shift in the timeline. The last term in \cref{eq:lif_update} becomes non-zero after each elicited spike when \(z_j^{t-1} = 1\) and partially resets the membrane voltage by the adaptive threshold \(v_j^{\text{th},t}\). In our N-MNIST experiments using \epropplus{}, the neuron model instead fully resets the membrane voltage to a fixed reset value. The adaptive threshold is updated as
    \begin{align}
      v_j^{\text{th},t} = v^\text{th} + \beta^\text{a} a_j^t.
    \end{align}
    The threshold adaptation evolves according to
    \begin{align}
      a_j^t = \rho a_j^{t-1} + z_j^{t-1},
      \label{eq:spike-threshold-adaptation}
    \end{align}
    where \( \rho = \exp\qty(-\frac{\Delta t}{\tau^\text{a}}) \), \( \beta^\text{a} \) is the prefactor of the threshold adaptation, and \( \tau^\text{a} \) is the adaptation time constant. LIF neurons without adaptation can be obtained by setting \( \beta^\text{a} \) to zero, yielding a constant spike threshold voltage \( v_j^{\text{th},t} \).

    Spikes from all recurrent neurons are integrated by the \(K\) output neurons modeled as leaky integrators:
    \begin{align}
      y_k^t & = \kappa y_k^{t-1} + \sum_{j=1}^J W_{kj}^\text{o} z_j^t,                    \\
            & = \sum_{j=1}^J W_{kj}^\text{o} \sum_{t'=1}^t \clash{\kappa}{t-t'} z_j^{t'},
      \label{eq:leaky_integrator}
    \end{align}
    using \cref{not:exponent-notation} and where \( \kappa = \exp \qty(-\frac{\Delta t}{\tau^\text{m,out}}) \) represents the decay factor of the membrane voltage over time, with \(\tau^\text{m,out}\) denoting the membrane time constant. The role of the output layer is to integrate and filter the recurrent activity, and convert it into a continuous output signal.

  \subsection{Loss function}
    Given a loss function \( \loss \) associated with a specific task, e-prop defines the error signal as the gradient of the loss with respect to the membrane voltage of the output neuron:
    \begin{align}
      E_k^t = \pdv{\loss}{y_k^t}.
      \label{eq:error_signal}
    \end{align}
    The scheme computes this error signal locally at the output neuron. Thereby, \( E_k^t\) is a direct measure of how the activity of the output neuron contributes to the overall loss.

    In regression tasks, the network learns to approximate the target signals \(y_k^{*,t}\) by the output signals \(y_k^t\). To quantify learning success, the original algorithm uses the mean squared error
    \begin{align}
      \loss & = \frac{1}{2} \sum_{t=1}^T \sum_{k=1}^K {\qty(y_k^t - y_k^{*,t})}^2,
      \label{eq:mean-squared-error}
    \end{align}
    and the corresponding error signals are
    \begin{align}
      E_k^t & = y_k^t - y_k^{*,t}.
    \end{align}
    In classification tasks, the original implementation computes the output signals \(\pi_k^t\) as probabilities using the softmax function
    \begin{align}
      \pi_k^t = \frac{\exp\qty(y_k^t)}{\sum_{k'=1}^K \exp\qty(y_{k'}^t)}.
      \label{eq:softmax}
    \end{align}
    The classifier is trained on the one-hot encoded target vector \(\pi^{*,t} \in {\{0,1\}}^K\) using the cross-entropy loss:
    \begin{align}
      \loss = -\sum_{t=1}^T \sum_{k=1}^K \pi_k^{*,t} \log \pi_k^t,
      \label{eq:cross-entropy-loss}
    \end{align}
    with the corresponding error signals
    \begin{align}
      E_k^t & = \sum_{k'=1}^K \pdv{\loss}{\pi_{k'}^t} \pdv{\pi_{k'}^t}{y_k^t} = \pi_k^t - \pi_k^{*,t}.
    \end{align}
    In classification tasks with \epropplus{}, in contrast, we employ a temporal mean squared error
    \begin{align}
      \loss = \frac{1}{K} \sum_{t=1}^T \sum_{k=1}^K {\qty(y_k^t-y_k^{*,t})}^2.
      \label{eq:temporal-mse}
    \end{align}
    to train the network on the one-hot encoded target vector \(y^{*,t} \in {\{0,1\}}^K\).

  \subsection{Gradients}\label{sec:gradients}
    This section summarizes some of the key steps involved in deriving the gradients for the original e-prop. Bellec et al. (2020) \supercite{bellec2020solution} showed that, at each step, the gradient of the loss \( \loss \) with respect to the weight \( W_{ji} \) factorizes into two terms,
    \begin{align}
      \dv{\loss}{W_{ji}} & = \sum_{t=1}^T \dv{\loss}{z_j^t} \underbrace{\dv{z_j^t}{W_{ji}}}_{\eqcolon e_{ji}^t}.
      \label{eq:original-recurrent-weight-update}
    \end{align}
    The second term, the eligibility trace \(e_{ji}^t\), captures local contributions. The first term accounts for all direct and indirect effects of the spike variable \( z_j^t \) on the loss, including influences through future steps. Computing this signal exactly requires
    BPTT\@. The e-prop scheme approximates this term by the partial derivative capturing only the direct dependence of the loss on the spike variable \supercite{bellec2020solution}, which, together with \cref{eq:error_signal,eq:leaky_integrator}, yields
    \begin{align}
      \dv{\loss}{z_j^{t'}} \approx \pdv{\loss}{z_j^{t'}} = \sum_{k=1}^K \pdv{\loss}{y_k^{t'}}\pdv{y_k^{t'}}{z_j^{t'}} = \sum_{k=1}^K W_{kj}^\text{o} \sum_{t=t'}^T E_k^{t} \clash{\kappa}{t-t'},
      \label{eq:pdv_L_z}
    \end{align}
    where we use \(t'\) instead of \(t\) for later convenience and \cref{not:exponent-notation}. This partial derivative can be computed locally and thus online. Substituting this into the gradient yields
    \begin{align}
      \dv{\loss}{W_{ji}}
       & \approx \sum_{t'=1}^{T} \pdv{\loss}{z_j^{t'}} e_{ji}^{t'}                                                                           \\
       & = \sum_{t'=1}^T \sum_{k=1}^K W_{kj}^\text{o} \sum_{t=t'}^T E_k^{t} \clash{\kappa}{t-t'} e_{ji}^{t'} \label{eq:summation-future}     \\
       & = \sum_{t=1}^T \sum_{k=1}^K W_{kj}^\text{o} E_k^t \sum_{t'=1}^t \clash{\kappa}{t-t'} e_{ji}^{t'} \label{eq:remove-summation-future} \\
       & = \sum_{t=1}^T \sum_{k=1}^K W_{kj}^\text{o} E_k^t \filter{F}{\kappa}{e_{ji}^t} \label{eq:filter-eligibility-trace}                  \\
       & \approx \sum_{t=1}^T \sum_{k=1}^K B_{jk} E_k^t \filter{F}{\kappa}{e_{ji}^t} \label{eq:replace-wkj-by-bjk}                           \\
       & = \sum_{t=1}^T L_j^t \filter{F}{\kappa}{e_{ji}^t}. \label{eq:final-gradient}
    \end{align}
    \cref{eq:remove-summation-future} uses the identity
    \begin{align}
      \sum_{t'=1}^T \sum_{t=t'}^T A^{t',t} = \sum_{t=1}^T \sum_{t'=1}^t A^{t',t}.
      \label{eq:index-flipping}
    \end{align}
    for an arbitrary matrix \(A\) to remove the summation over future errors by interchanging the summation indices, allowing the algorithm to operate online.  Effectively swapping row- and column-wise summations, this transforms summing over all future times \( t \geq t' \) for each past time \(t'\) into summing over all past times \( t' \leq t \) for current time \(t\). The final term becomes a low-pass filtered version of the eligibility trace.\ \cref{eq:filter-eligibility-trace} introduces a shorthand for filtering a dynamic variable \(u_i\) with constant \(\gamma \) by
    \begin{align}
      \filter{F}{\gamma}{u_i^t} & = \gamma \filter{F}{\gamma}{u_i^{t-1}} + u_i^t,\quad\filter{F}{\gamma}{u_i^0}=0.
      \label{eq:filter}
    \end{align}

    \cref{eq:replace-wkj-by-bjk} replaces the output weight matrix \(W_{kj}^\text{o}\) by the feedback weight matrix \(B_{jk}\), which in BPTT is symmetric to the output weights, i.e., \( B_{jk} = {W_{kj}^\text{o}}^\top \). E-prop avoids this biologically implausible symmetry by assigning fixed random values to \(B_{jk}\), following the idea of feedback alignment \supercite{lillicrap2016random}, where output weights adapt during training to align with random feedback weights.
    In direct feedback alignment \supercite{nokland2016direct}, the output layer sends the error signal through a distinct random feedback matrix to each hidden layer, whereas in broadcast alignment \supercite{samadi2017deep} it uses the same random feedback matrix for all hidden layers. In single-layer networks as used in all experiments here, these methods are equivalent. The error signals multiplied by the feedback weight matrix and summed over all output neurons yield the learning signals \(L_j^t\) in \cref{eq:final-gradient}. In this final expression, the eligibility trace and the learning signal are multiplied at each step of \SI{1}{\milli\second}, and the resulting products are summed over a sample of typically \SIrange{1}{2}{\second}.

    The eligibility trace can be calculated as
    \begin{align}
      e_{ji}^t = \dv{z_j^t}{W_{ji}} & = \pdv{z_j^t}{\vb*{h}_j^t} \underbrace{\sum_{t'=1}^t \pdv{\vb*{h}_j^t}{\vb*{h}_j^{t-1}} \cdots \pdv{\vb*{h}_j^{t'+1}}{\vb*{h}_j^{t'}} \cdot \pdv{\vb*{h}_j^{t'}}{W_{ji}}}_{\eqcolon \vb*{\epsilon}_{ji}^t},
      \label{eq:eligibility-trace}
    \end{align}
    where
    \begin{align}
      \vb*{h}_j^t & = \mqty(v_j^t \\ a_j^t)
    \end{align}
    represents the hidden state variables of the recurrent neurons and \(\vb*{\epsilon}_{ji}^t\) is the eligibility vector. The gradient before the eligibility vector in \cref{eq:eligibility-trace}  represents the postsynaptic information:
    \begin{align}
      \pdv{z_j^t}{\vb*{h}_j^t} & = {\mqty(\pdv{z_j^t}{v_j^t} \\ \pdv{z_j^t}{a_j^t})}^\top \approx {\mqty(\psi_j^t \\ -\beta^\text{a} \psi_j^t)}^\top.
    \end{align}
    The surrogate gradient (or pseudo-derivative function) \( \psi_j^t \) captures the relationship between the non-differentiable discrete postsynaptic spike state variable and the postsynaptic membrane voltage. The original e-prop scheme uses the piecewise-linear surrogate gradient:
    \begin{align}
      \psi_j^t & = \gamma \max\qty(0, 1 - \beta \abs{v_j^t - v_j^{\text{th},t}}),
      \label{eq:piecewise-linear-surrogate-gradient}
    \end{align}
    which peaks with a magnitude of the prefactor \( \gamma \) at the spike threshold, \( v_j^{\text{th},t} \), and linearly decreases to zero in both positive and negative directions. Compared to the original definition \supercite{bellec2020solution}, we redefine the prefactor as \( \gamma = \frac{\gamma_\text{original}}{v_{\text{th}^t}} \), introducing a scaling factor \( \beta \) to incorporate \( \frac{1}{v^{\text{th},t}} \) from the original definition.
    For \epropplus{}, we replace this function by the exponential surrogate gradient\supercite{shrestha2018slayer}
    \begin{align}
      \psi_j^t = \gamma \exp\qty(-\beta \abs{v_j^t-v_j^{\text{th},t}}).
      \label{eq:exponential-surrogate-gradient}
    \end{align}

    The eligibility vector \( \vb*{\epsilon}_{ji}^t \) includes presynaptic information and is computed recursively from the gradients of the hidden state variables as
    \begin{align}
      \vb*{\epsilon}_{ji}^t & = \pdv{\vb*{h}_j^t}{\vb*{h}_j^{t-1}} \cdot \vb*{\epsilon}_{ji}^{t-1} + \pdv{\vb*{h}_j^t}{W_{ji}}.
      \label{eq:eligibility-vector}
    \end{align}
    The gradients are given by
    \begin{align}
      \pdv{\vb*{h}_j^t}{\vb*{h}_j^{t-1}} & = \mqty(\pdv{v_j^t}{v_j^{t-1}}         & \pdv{v_j^t}{a_j^{t-1}} \\
      \pdv{a_j^t}{v_j^{t-1}}             & \pdv{a_j^t}{a_j^{t-1}}) = \mqty(\alpha & 0                      \\
      \psi_j^{t-1}                       & \rho - \psi_j^{t-1} \beta^\text{a}),                            \\
      \pdv{\vb*{h}_j^t}{W_{ji}}          & = \mqty(\pdv{v_j^t}{W_{ji}}                                     \\
      \pdv{a_j^t}{W_{ji}}) = \mqty(z_i^{t-1}                                                               \\
      0).
    \end{align}
    where \( \alpha \) denotes the decay factor of the membrane voltage as introduced in \cref{eq:lif_update}. Substituting all gradients into \cref{eq:eligibility-vector} yields
    \begin{align}
      \vb*{\epsilon}_{ji}^t & = \mqty(\filter{F}{\alpha}{z_i^{t-1}} \\ \psi_j^{t-1} \epsilon_{ji}^{\text{v},t-1} + \qty(\rho - \psi_j^{t-1} \beta^\text{a}) \epsilon_{ji}^{\text{a},t-1}),
      \label{eq:eligibility-vector-substituted}
    \end{align}
    where the subscripts \(\text{v}\) and \(\text{a}\) denote the components of the eligibility vector associated with the membrane potential and the adaptation variable, respectively.

    Using this, the full gradient is expressed as
    \begin{align}
      \dv{\loss}{W_{ji}} & \approx \sum_{t=1}^T L_j^t \filter{F}{\kappa}{\psi_j^t\qty(\filter{F}{\alpha}{z_i^{t-1}} - \beta^\text{a} \epsilon_{ji}^{\text{a},t})}.
      \label{eq:filtered-eligibility-trace}
    \end{align}
    The equations derived so far apply only to the recurrent weights, yielding the gradient \(g_{ji}^{\text{r},t}\), while analogous calculations provide \( g_{ji}^{\text{i},t}\) and \(g_{kj}^{\text{o},t}\) for the input and output weights, respectively:
    \begin{align}
      g_{ji}^{\text{r},t} & = L_j^t \filter{F}{\kappa}{\psi_j^t \filter{F}{\alpha}{z_i^{t-1}}- \beta^\text{a} \epsilon_{ji}^{\text{a},t}}, \label{eq:gradient-rec} \\
      g_{ji}^{\text{i},t} & = L_j^t \filter{F}{\kappa}{\psi_j^t \filter{F}{\alpha}{x_i^t}- \beta^\text{a} \epsilon_{ji}^{\text{a},t}}, \label{eq:gradient-in}      \\
      g_{kj}^{\text{o},t} & = E_k^t.
      \label{eq:gradient-out}
    \end{align}
    The calculation of the gradients for output synapses is simpler because it does not involve a surrogate gradient, as the output neurons are modeled as leaky integrators without a spiking mechanism.

  \subsection{Firing rate regularization}

    Firing rate regularization introduces a penalty when \( f_j \), the firing rate of recurrent neuron \( j \) averaged over the steps \( T \) of a sample, deviates from a set target firing rate \(f^*\):
    \begin{align}
      \loss^\text{reg} & = \frac{c^\text{reg}}{2} \sum_{j=1}^J {\qty(f_j - f^*)}^2, \\
      f_j              & = \frac{1}{T} \sum_{t=1}^T z_j^t.
      \label{eq:loss-firing-rate-regularization}
    \end{align}
    For notational simplicity, the firing rate is defined as spikes per \SI{1000}{} steps, which is equivalent to
    \SI{}{\spikepersecond}. This mechanism ensures that throughout the optimization, the firing rate of each recurrent neuron stays close to a desired target firing rate despite the weight updates. It is realized by adding
    \begin{align}
      g_{ji}^{\text{reg},t} & = \dv{\loss^\text{reg}}{W_{ji}} = \frac{c^\text{reg}}{T}\qty(f_j-f^*)e_{ji}^t.
      \label{eq:firing-rate-regularization}
    \end{align}
    to the recurrent gradient \(g_{ji}^\text{r}\). The term is negative if the average firing rate is larger than the target firing rate, thus decreasing the weight, and positive if it is smaller, thus increasing the weight. While the deviation is specific to the postsynaptic neuron, each synapse multiplies the deviation with its eligibility trace.

    In contrast to this static firing rate regularization, where the firing rate is calculated from the total number of spikes in a sample, the original scheme additionally defines a dynamic variant in which the regularization loss depends on a firing rate that varies over the course of the sample:
    \begin{align}
      \loss^\text{reg} & =  \frac{c^\text{reg}}{2} \sum_{t=1}^T \sum_{j=1}^J {\qty(f_j^t - f^*)}^2, \label{eq:sma-loss}        \\
      f_j^t            & = \frac{1}{t}\sum_{t'=1}^t z_j^{t'}  = \beta^t f_j^{t-1} + \qty(1-\beta^t) z_j^t, \label{eq:sma-rate}
    \end{align}
    where \( \beta^t = \frac{t}{t+1} \) and \( c^\text{reg} \) is the regularization coefficient.
    For \epropplus{}, we use an exponential moving average by replacing \(\beta^t\) with a constant \( \beta \), thus converting the operation into a low-pass filter \( \filter{G}{}{\cdot} \):
    \begin{align}
      f_j^t
      = \beta f_j^{t-1} + \qty(1 - \beta) z_j^t
      = \qty(1-\beta) \sum_{t'=1}^t \clash{\beta}{t-t'} z_j^{t'}
      \eqcolon \filter{G}{\beta}{z_j^t}.
      \label{eq:ema-rate}
    \end{align}
    using \cref{not:exponent-notation}.
    By applying the e-prop principle \supercite{bellec2020solution} of considering only local interactions and the eligibility trace definition, we have
    \begin{align}
      \dv{f_j^t}{W_{ji}} & = \dv{\filter{G}{\beta}{z_j^t}}{W_{ji}}
      = \qty(1-\beta) \sum_{t'=1}^t \clash{\beta}{t-t'} \dv{z_j^{t'}}{W_{ji}}
      = \filter{G}{\beta}{\dv{z_j^t}{W_{ji}}}
      \approx \filter{G}{\beta}{e_{ji}^t},
    \end{align}
    which yields the gradient of the loss associated with dynamic firing rate regularization
    \begin{align}
      g_{ji}^{\text{reg},t} & = c^\text{reg} \qty( f_j^t - f^*) \dv{f_j^t}{W_{ji}}
      \approx c^\text{reg}\qty(f_j^t - f^*) \filter{G}{\beta}{e_{ji}^t}.
      \label{eq:ema-gradient}
    \end{align}

  \subsection{Optimization}
    The overall gradient is computed from the accumulated per-step gradients:
    \begin{align}
      \loss(W_{ji})                                                      & = \sum_{t=1}^T \ell^t\!\qty(W_{ji}) \,,                                              \\
      \underbrace{\pdv{\loss}{W_{ji}} \qty(W_{ji}^n)}_{\eqqcolon g_{ji}} & = \sum_{t=1}^T \underbrace{\pdv{\ell^t}{W_{ji}}\qty(W_{ji}^n)}_{\eqqcolon g_{ji}^t}.
      \label{eq:accumulated-gradients}
    \end{align}
    The new weight at the end of each sample then results from
    \begin{align}
      W_{ji}^{n+1} & = W_{ji}^n + \Delta W_{ji},
    \end{align}
    where \(\Delta W_{ji}\) denotes the weight update computed in a single optimization step. In the case of gradient descent, the weight update is
    \begin{align}
      \Delta W_{ji} = -\eta \sum_{t=1}^T g_{ji}^t,
      \label{eq:gradient-descent}
    \end{align}
    where \( \eta \) is the learning rate. A more robust and typically faster-converging gradient-based optimization method is the Adam algorithm \supercite{kingma2017adam}. Unlike gradient descent, this update cannot be applied directly to the sum of gradients, because two internal variables, \( m_{ji}^ t\) and \( v_{ji}^t \), depend on the gradient at each step and must be updated sequentially:
    \begin{align}
      m_{ji}^0       & = 0, \quad v_{ji}^0 = 0,                                     \\
      m_{ji}^t       & = \beta_1 m_{ji}^{t-1} + \qty(1-\beta_1) g_{ji}^t,           \\
      v_{ji}^t       & = \beta_2 v_{ji}^{t-1} + \qty(1-\beta_2) {\qty(g_{ji}^t)}^2, \\
      \hat{m}_{ji}^t & = \frac{m_{ji}^t}{1-\clash{\beta_1}{t}}, \label{eq:adam1}    \\
      \hat{v}_{ji}^t & = \frac{v_{ji}^t}{1-\clash{\beta_2}{t}}, \label{eq:adam2}
    \end{align}
    where typical values for the exponential decay rates are \( \beta_1=0.9 \) and \( \beta_2=0.999\) and using \cref{not:exponent-notation}. After accumulating the gradient contributions over all time steps  \(t = 1, \dots, T\), the Adam optimizer performs a single weight update at the end of the sequence according to
    \begin{align}
      \Delta W_{ji} & = - \eta \sum_{t=1}^T \frac{\hat{m}_{ji}^t}{\sqrt{\hat{v}_{ji}^t} + \epsilon}, \label{eq:adam}
    \end{align}
    where the small numerical stabilization constant is usually set to \( \epsilon=10^{-8}\).
    The first moment estimate \( m_{ji}^t \) corresponds to the exponential moving average of past gradients (mean of past gradients), and the second moment estimate \( v_{ji}^t \) corresponds to the exponential moving average of squared gradients (variance of past gradients). The algorithm corrects both variables for the initialization bias caused by starting at zero (\cref{eq:adam1,eq:adam2}). For comparability with time-driven e-prop \supercite{bellec2020solution}, our implementation follows TensorFlow \supercite{abadi2016tensorflow}, which reorders the computations in \cref{eq:adam1,eq:adam2,eq:adam} as proposed in the original Adam algorithm \supercite{kingma2017adam}:
    \begin{align}
      \eta^t        & = \eta \frac{\sqrt{1-\clash{\beta_2}{t}}}{1-\clash{\beta_1}{t}}, \label{eq:eta-adam-fast} \\
      \Delta W_{ji} & = - \sum_{t=1}^T \eta^t \frac{m_{ji}^t}{\sqrt{v_{ji}^t} + \hat{\epsilon}}.
      \label{eq:delta-w-adam-fast}
    \end{align}
    Moreover, we follow TensorFlow in assuming a constant \(\hat{\epsilon}\) with default value \(10^{-7}\) in the expression \(\hat{\epsilon} = \epsilon \sqrt{ 1 - \clash{\beta_2}{t} }\), whereas the original Adam algorithm fixes \(\epsilon \).

    %\subsection{Biological connectivity and weights}\label{sec:bio-conn-weights}
    %  In ANNs, weights can be positive or negative. In contrast, biological neurons release only one type of neurotransmitter, so under typical conditions a neuron's outgoing synapses are exclusively excitatory (positive) or inhibitory (negative)---a principle known as Dale's law \supercite{strata1999dale}. While synapse type is fixed, strength can change (functional plasticity), and synapses can disappear or form anew (structural plasticity). In the primate cortex, the ratio of excitatory to inhibitory neurons is about 4:1, with a local connection probability of roughly \SI{0.1}{}. Imposing these biological constraints on an architecture trained with time-driven e-prop together with a stochastic rewiring algorithm \supercite{bellec2018deep} was shown to preserve performance on the evidence-accumulation task (see Fig. 3c in Bellec et al., 2020 \supercite{bellec2020solution}).

  \subsection{Tasks}

    \subsubsection{Pattern generation}\label{sec:pattern-generation}
      In this regression task, the network learns to generate a signal that is one second in duration and composed of the summation of four sinusoids, each with randomly assigned phases and amplitudes. After training, the network approximates the target signal given the specific frozen spike input pattern. The network output is projected on one output neuron whose membrane voltage \(y_k\) fluctuates around zero before training and follows the target signal after training. Results from experiments on the pattern generation task are shown in \cref{fig:accuracy_comparison_event_time_driven-a,fig:accuracy_comparison_event_time_driven-b,fig:proof_of_concept_regression,fig:additional-tasks,fig:etrace-filter}.

    \subsubsection{Evidence accumulation}\label{sec:evidence-accumulation}
      The evidence accumulation task is a classification problem inspired by a behavioral task in which a mouse runs on a linear track and receives cues on the left and right. At the end of the track, it has to decide whether to turn left or right. The network must learn from the data that the correct choice corresponds to the side with the most cues. In the spiking network, two input populations provide Poisson spike trains that represent the cues. A third input population provides background input throughout the task, and a fourth is only active at the end of a sample, indicating the phase when the network must decide. The plasticity is turned on only in this last period, so we refer to it as the learning window. A long intermediate phase between the presentation of the cues and the onset of the recall phase with solely background input adds an extra challenge to this task since the network needs to keep the cues in memory. Following the original model\supercite{bellec2020solution}, the network's capacity to memorize the cues arises from the slowly decaying spike threshold adaptation (\cref{eq:spike-threshold-adaptation}) of the recurrent LIF neurons (\cref{eq:lif_update}). Alternatively, increasing the membrane time constant\supercite{frenkel2022reckon} also permits memory to span longer timescales. Results from experiments on the evidence accumulation task are shown in \cref{fig:accuracy_comparison_event_time_driven-c,fig:accuracy_comparison_event_time_driven-d,fig:proof_of_concept_classification}.

    \subsubsection{Neuromorphic MNIST}\label{sec:neuromorphic-mnist}
      The N-MNIST dataset \supercite{orchard2015converting}, an adaptation of MNIST for handwritten digits designed for neuromorphic computing, was created by converting the MNIST computer vision dataset into a dataset of spike sequences using dynamic vision sensors. In these sensors, each pixel responds only to changes in the scene caused by variations in brightness or motion. The 2D MNIST images were displayed sequentially on a monitor, while a sensor mounted on a pan-tilt camera platform scanned them in a plane parallel to the screen, ensuring independence from scene depth. Using a real sensor rather than a simulation introduces sensor noise, reflecting real-world conditions. Moving the sensor instead of the scene is a biologically plausible sensing approach. The camera performed three 100 ms micro-saccades, tracing an isosceles triangle (upper left to lower middle to upper right to upper left), mimicking small subconscious eye movements. Each dataset image comprises a 34 \( \times \) 34 pixel grid and is represented as a list of binary events. These events are characterized by a timestamp, x and y pixel coordinates, and a binary change in pixel intensity (0 for a decrease, 1 for an increase). Pixels with unchanged intensities are not recorded. These binary events can be interpreted as spike trains, mimicking biological neural processing, making the dataset particularly well-suited for SNNs trained with the e-prop algorithm. To enhance computational efficiency, we exclude pixels that generate no or very few events. We represent each remaining pixel by a spike generator, which emits a spike for every ON event registered in that pixel. Each spike generator sends these spikes to a corresponding input neuron. The input neurons project onto a recurrent network, which is further connected to 10 output neurons --- one for each digit class. Each output neuron compares the network signal to the teacher signal representing the correct digit class. Results from experiments on the N-MNIST task are shown in \cref{fig:accuracy_comparison_with_without_bio,fig:nmnist-bio-traces,fig:nmnist-original-traces,fig:scaling,fig:surrogate-gradient-functions}.

  \subsection{Scaling}\label{sec:scaling-methods}

    We measure the time required for state propagation, excluding the time needed for network construction and analysis. First, we conduct experiments for a strong scaling scenario, where the model size remains fixed while the compute system is scaled up. Second, we perform experiments for a weak scaling scenario, where the model size increases proportionally with the size of the compute system.

    To avoid the risk of the network entering a high-activity or high-synchrony state\supercite{albada2015scalability}, which could make comparisons unfair, we use an ignore-and-fire mechanism for all scaling experiments. In this approach, neurons ignore their inputs and spike randomly at a constant firing rate, following an established mechanism \supercite{espinoza2024nest}. Since neurons in the brain are not fully connected (where the number of synapses would increase quadratically with network size), we consider a more biologically realistic scenario: as the number of neurons increases, the number of synapses per neuron is kept constant at a biologically plausible in-degree. As a side effect, this ensures that the firing rate remains approximately constant \supercite{jordan2018extremely,lansner2012virtues} but correlations decrease.

    In this setting, we evaluate weak (see \cref{fig:scaling-a}) and strong scaling (see \cref{fig:scaling-b}) using networks of ignore-and-fire neuron models firing at randomized phases at a rate of \SI{5}{\spikepersecond}. The base network contains \SI{128000}{} neurons. The base input layer consists of \SI{1000}{} spike generators firing with Poisson statistics and a rate of \SI{5}{\spikepersecond}. The base output layer consists of \SI{10}{} neurons. Input connections have an indegree of \SI{100}{}, recurrent connections \SI{10000}{}, output connections \SI{1000}{}, and feedback connections an outdegree of \SI{100}{}. Recurrent connections exclude autapses and multapses --- self-connections and multiple synapses between the same pair of neurons \supercite{senk2022connectivity} --- and are plastic, except in the static case. In weak scaling, the input, recurrent, and output layers are proportionally scaled. For e-prop and \epropplus{}, each output neuron receives a target signal from a target generator and broadcasts it to all recurrent neurons. Simulations cover \SI{20}{\second} of biological time resulting in \SI{20000}{} steps of \SI{1}{\milli\second} each. Each simulation uses \SI{32}{} CPU cores per task and \SI{4}{} tasks per node. Statistical results are based on \SI{5}{} runs with different random seeds.

\section*{Acknowledgments}
  We thank Wolfgang Maass for raising the question of whether a scalable implementation of e-prop for sparse networks could be found. Jakob Jordan and Alexander van Meegen implemented an early event-driven offline e-prop algorithm in NEST (unpublished).\ \cref{fig:additional-tasks} is the result of joint work with Charl Linssen, inspired by activities and feedback at the CapoCaccia Workshop toward Neuromorphic Intelligence 2023. The project also benefited from discussions with Franz Scherr on the original e-prop model, review comments on the NEST implementation from Charl Linssen, feedback on the documentation from Jessica Mitchell, and technical assistance from Dennis Terhorst.

\section*{Funding}
  This work was supported by Joint Lab ``Supercomputing and Modeling for the Human Brain'' (SMHB); HiRSE\_PS\@; NeuroSys (Clusters4Future, BMBF, 03ZU1106CB); EU Horizon 2020 Framework Programme for Research and Innovation (945539, Human Brain Project SGA3) and Horizon Europe Programme under the Specific Grant Agreement No. 101147319 (EBRAINS 2.0 Project); computing time granted by the JARA Vergabegremium and provided on the JARA Partition part of the supercomputer JURECA at Forschungszentrum Jülich (computation grant JINB33); and Käte Hamburger Kolleg: Cultures of Research (c:o/re), RWTH Aachen (BMBF, 01UK2104).
  % Open access publication funded by the Deutsche Forschungsgemeinschaft (DFG, German Research Foundation) – 491111487.

  \appendix
\part*{Supplementary Information}

  \section{Figures}
    \begin{suppfigure}[H]
      \centering
      \includegraphics[width=0.8\textwidth]{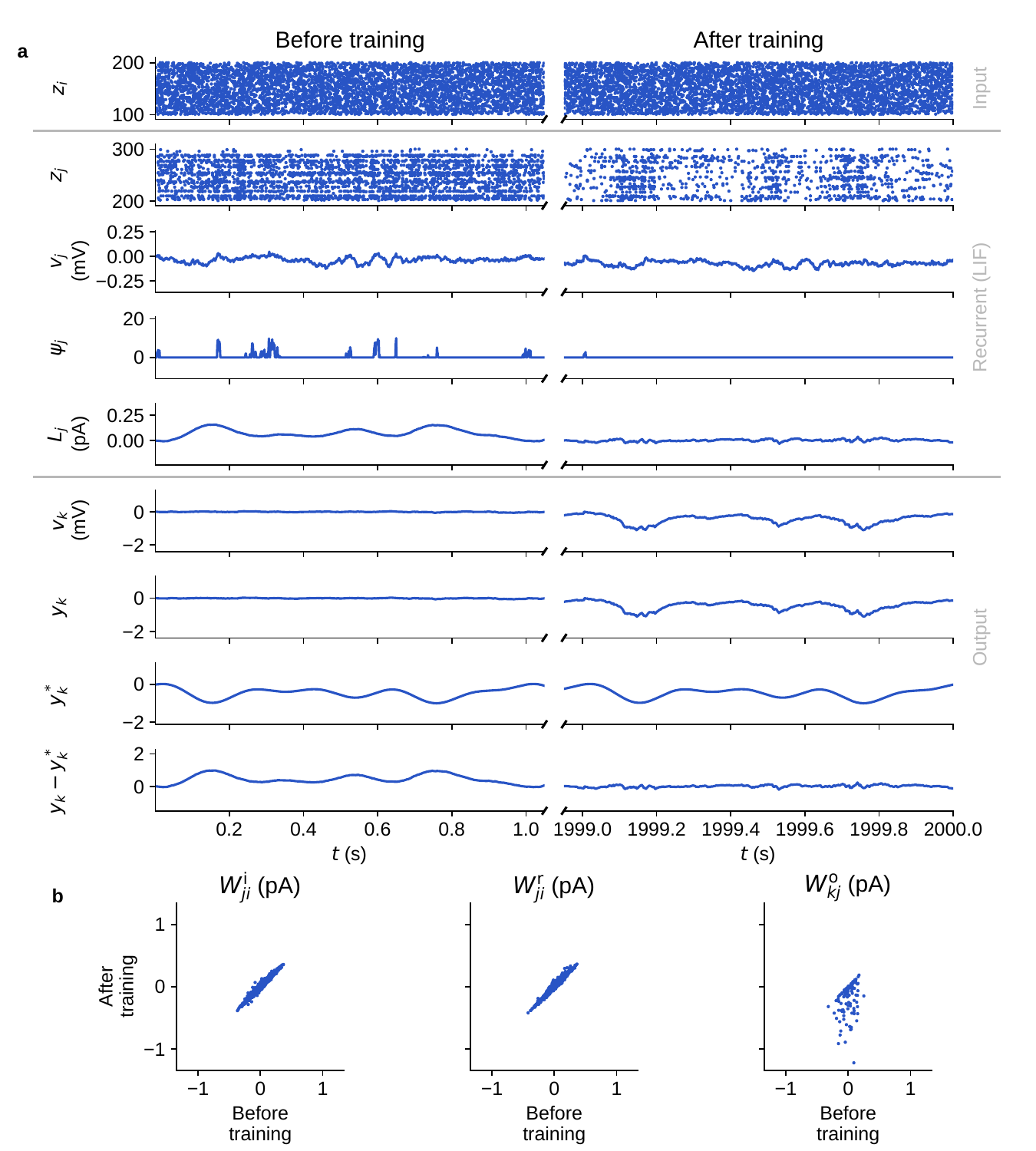}
      \phantomsubfigure{fig:proof_of_concept_regression-a}%
      \phantomsubfigure{fig:proof_of_concept_regression-b}%
      \caption[Pattern generation as a regression task with event-driven e-prop]{\captiontitle{Pattern generation as a regression task with event-driven e-prop.}
        \subfiglabel{a} Time traces of dynamic variables recorded before and after training.
        \subfiglabel{b} Distributions of input, recurrent, and output weights before vs.\ after training, indicating that the task can be solved with plasticity restricted to output synapses.
      }\label{fig:proof_of_concept_regression}
    \end{suppfigure}

    \begin{suppfigure}[H]
      \centering
      \includegraphics[width=0.8\textwidth]{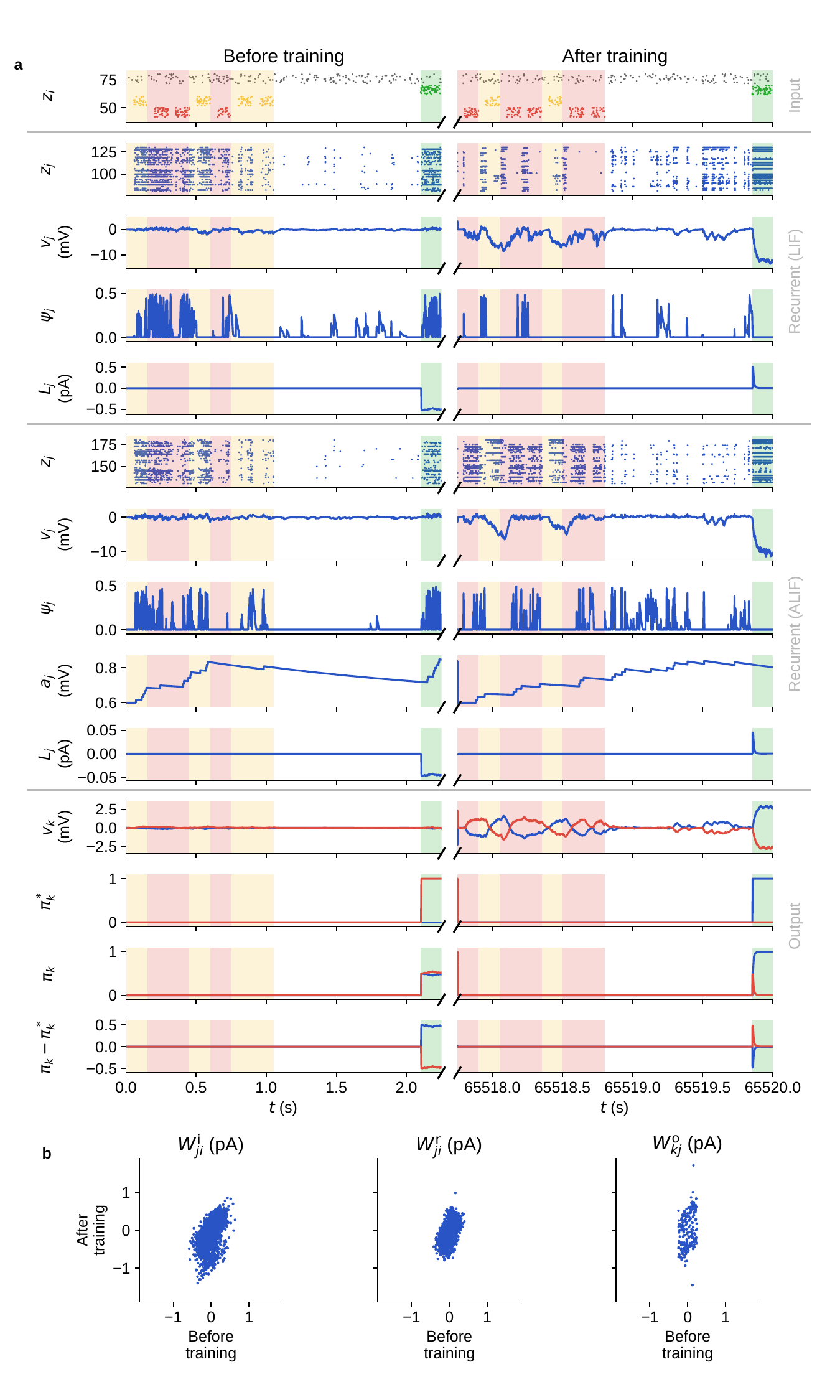}
      \phantomsubfigure{fig:proof_of_concept_classification-a}%
      \phantomsubfigure{fig:proof_of_concept_classification-b}%
      \caption[Evidence accumulation modeled as a classification task using event-driven e-prop]{\captiontitle{Evidence accumulation modeled as a classification task using event-driven e-prop.}
        \subfiglabel{a} Time traces of dynamic variables recorded before and after training. Yellow and red spikes represent the firing of two neuron populations corresponding to cues observed by a mouse in a behavioral experiment on the left and right sides while running along a linear track, respectively. Green spikes represent the firing of a population indicating the decision phase after a latency period.
        \subfiglabel{b} Distributions of input, recurrent, and output weights before vs.\ after training.
      }\label{fig:proof_of_concept_classification}
    \end{suppfigure}

    \begin{suppfigure}[H]
      \centering
      \includegraphics[width=\textwidth]{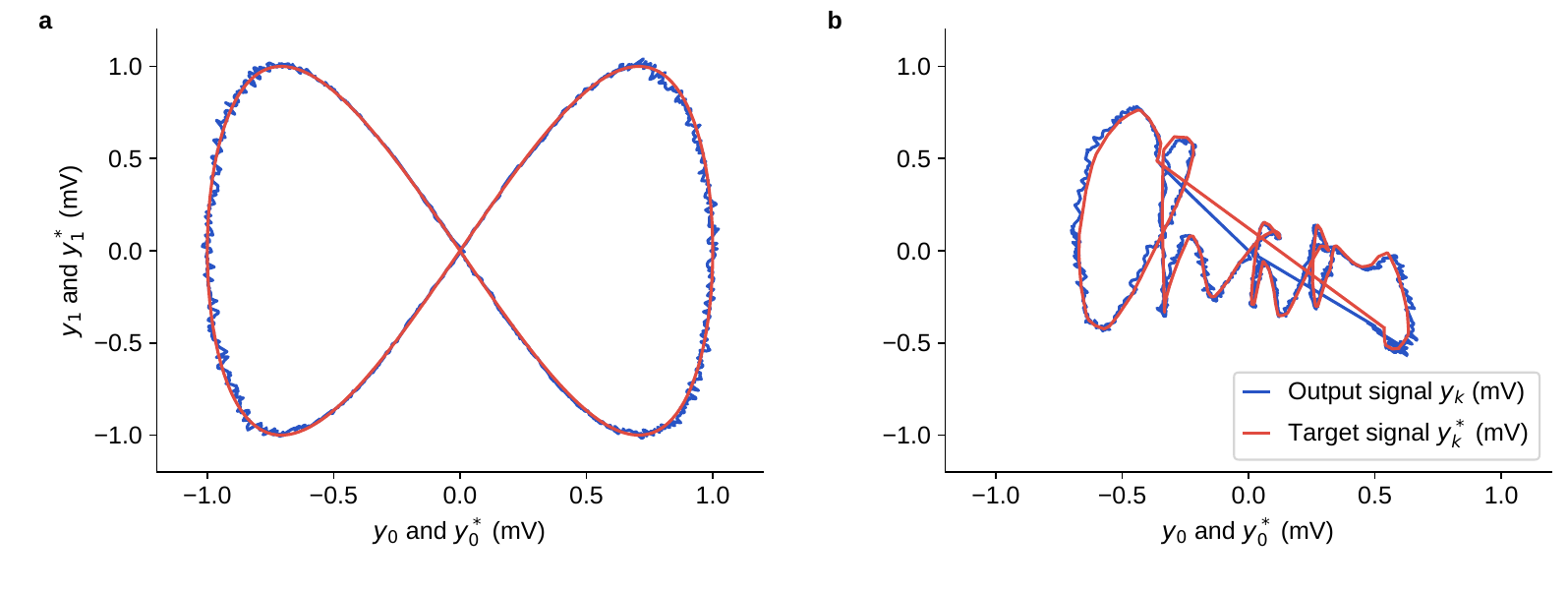}
      \phantomsubfigure{fig:additional-tasks-a}%
      \phantomsubfigure{fig:additional-tasks-b}%
      \caption[Generation of a lemniscate and handwriting pattern]{\captiontitle{Pattern generation with two output neurons.} A lemniscate pattern \subfiglabel{a} and the handwritten word `chaos' \subfiglabel{b}, with two output neurons encoding the horizontal and vertical coordinates, trained for \SI{4000}{} iterations.
      }\label{fig:additional-tasks}
    \end{suppfigure}

    \begin{suppfigure}[H]
      \centering
      \includegraphics[width=\textwidth]{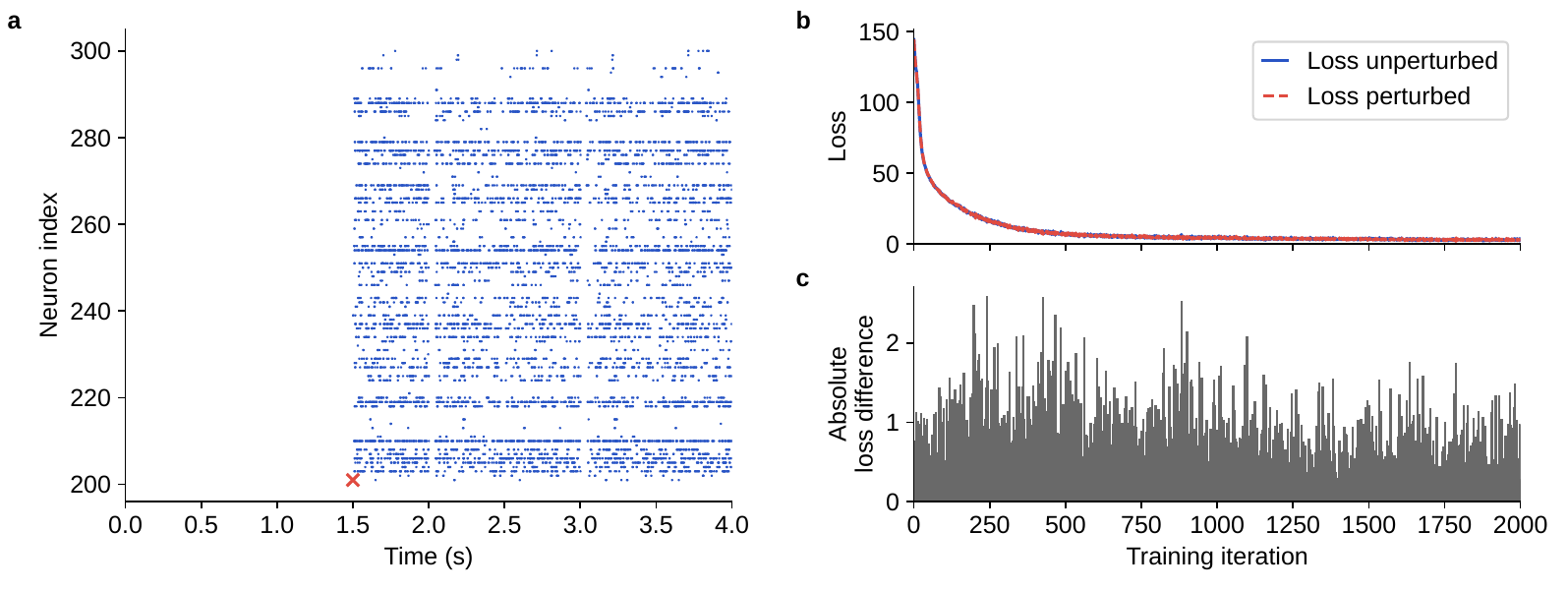}
      \phantomsubfigure{fig:spike-perturbation-a}%
      \phantomsubfigure{fig:spike-perturbation-b}%
      \phantomsubfigure{fig:spike-perturbation-c}%
      \caption[Impact of a perturbation spike on learning dynamics]{\captiontitle{Impact of a perturbation spike on learning dynamics during the regression task.}
        \subfiglabel{a} Comparison between two simulations, one with and one without a forced spike from neuron \SI{201}{} at \SI{1.5}{\second} (red cross). Deviations between the two activity patterns, binned into 1 ms intervals, correspond to bins containing a spike in only one simulation (shown as dots).
        \subfiglabel{b} Loss comparison between perturbed and unperturbed simulations.
        \subfiglabel{c} Absolute difference between the loss of the perturbed simulation and the loss of the unperturbed simulation, computed for each iteration individually.
      }\label{fig:spike-perturbation}
    \end{suppfigure}

    \begin{suppfigure}[H]
      \centering
      \includegraphics[width=\textwidth]{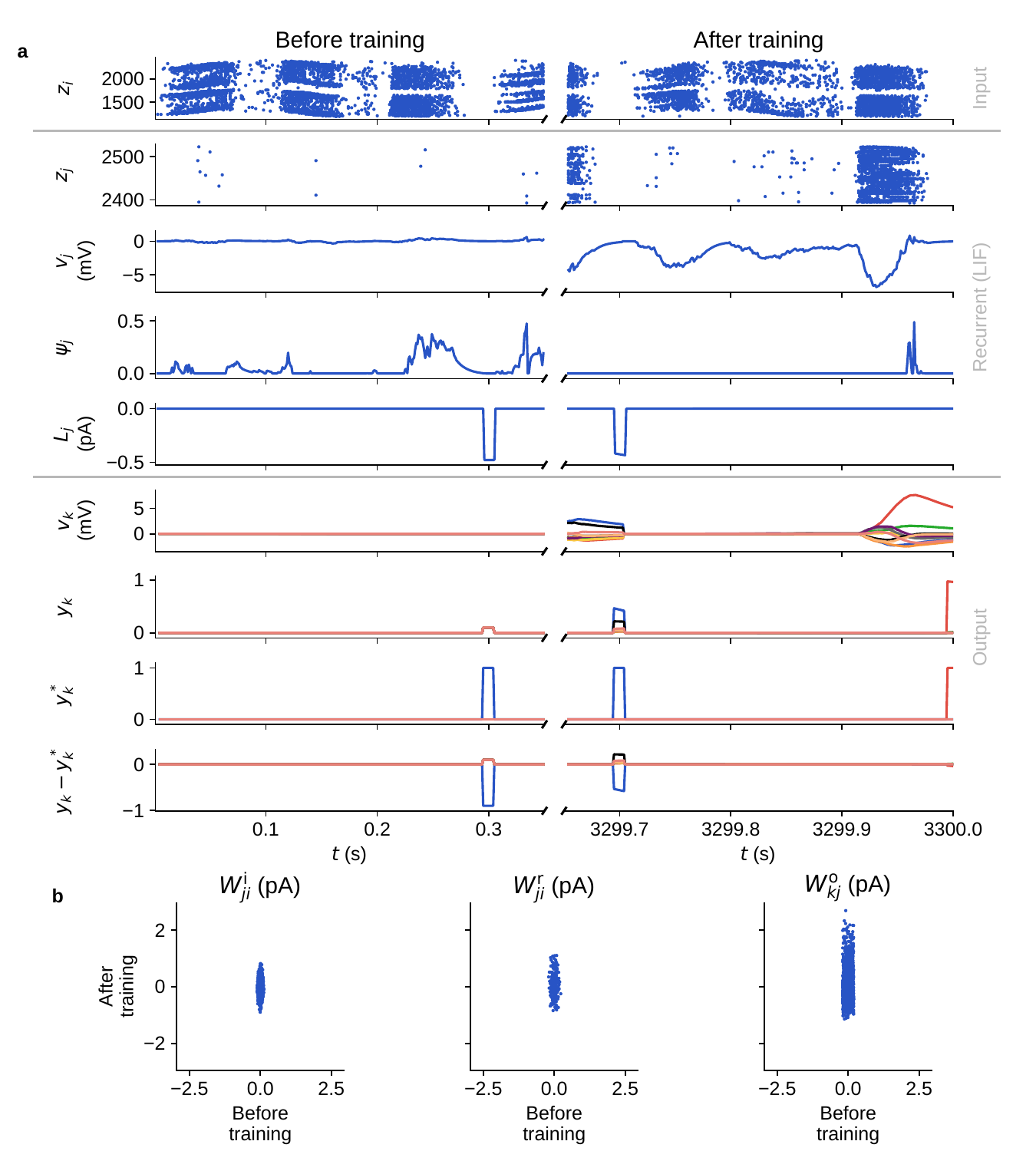}
      \phantomsubfigure{fig:nmnist-original-traces-a}%
      \phantomsubfigure{fig:nmnist-original-traces-b}%
      \caption[Dynamics and weight distributions before and after training N-MNIST using event-driven e-prop]{\captiontitle{Dynamics and weight distributions before and after training N-MNIST using event-driven e-prop.}
        \subfiglabel{a} Time traces of dynamic variables recorded before and after training.
        \subfiglabel{b} Distributions of input, recurrent, and output weights.
      }\label{fig:nmnist-original-traces}
    \end{suppfigure}

    \begin{suppfigure}[H]
      \centering
      \includegraphics[width=\textwidth]{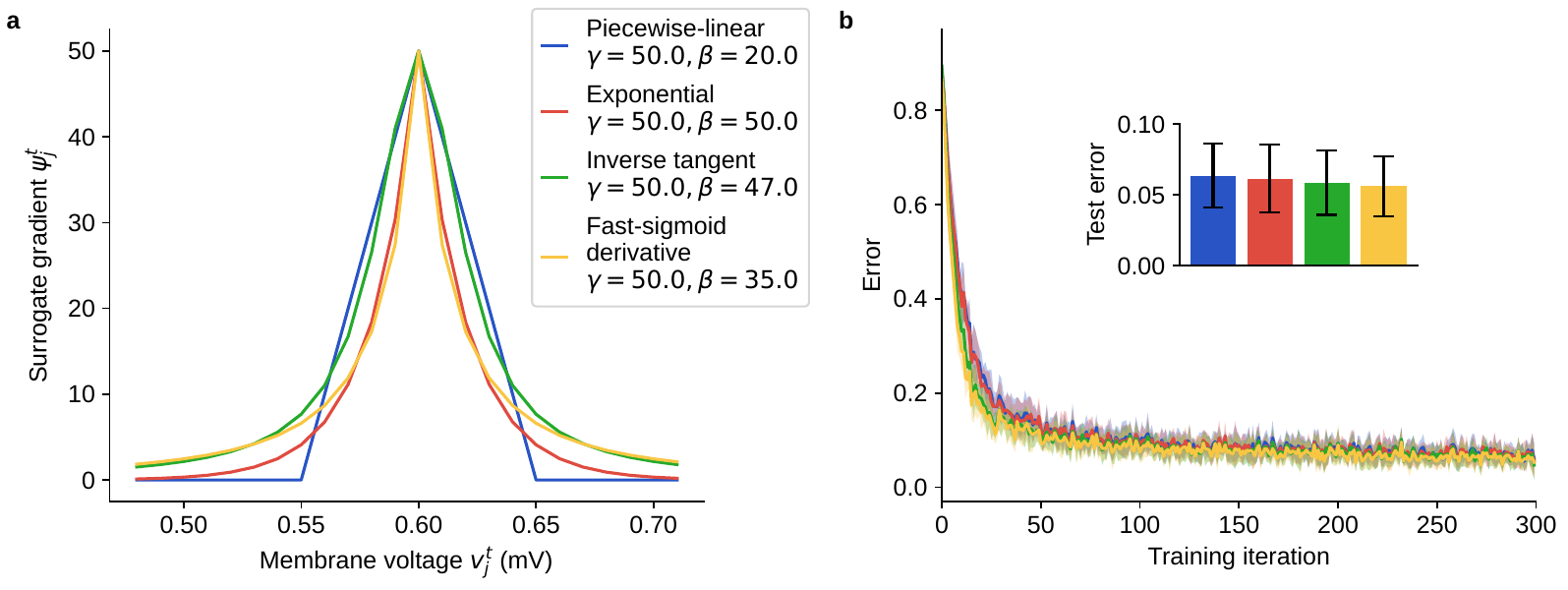}
      \phantomsubfigure{fig:surrogate-gradient-functions-a}%
      \phantomsubfigure{fig:surrogate-gradient-functions-b}%
      \caption[Learning performance comparison between different surrogate gradients]{\captiontitle{Learning performance comparison between different surrogate gradients.}
      \subfiglabel{a} Profiles of different surrogate gradient functions as a function of voltage, with a threshold voltage set at \SI{0.6}{\milli\volt}. Alongside the piecewise-linear (\cref{eq:piecewise-linear-surrogate-gradient}) and exponential (\cref{eq:exponential-surrogate-gradient}) functions, we also investigate a function corresponding to the derivative of a fast sigmoid \supercite{zenke2018superspike}, \( \psi_j^t = \gamma / {\qty(1 + \beta \abs{v_j^t-v_j^{\text{th},t}})}^2 \), and an inverse tangent function \supercite{fang2021deep} \(\psi_j^t = \gamma / \qty(1+{\qty(\beta{\qty(v_j^t-v_j^{\text{th},t})})}^2) \).
      \subfiglabel{b} Training error time courses for the N-MNIST task comparing models with surrogate gradients shown in (a). Curves represent the mean and shaded areas the standard deviation across \SI{10}{} trials with different random seeds. Bars in the inset show the mean across all trials and \SI{10}{} test iterations per trial, and error bars represent the combined standard deviation, calculated as the square root of the sum of within-trial and between-trial variances.}\label{fig:surrogate-gradient-functions}
    \end{suppfigure}

    \begin{suppfigure}[H]
      \centering
      \includegraphics[width=0.6\textwidth]{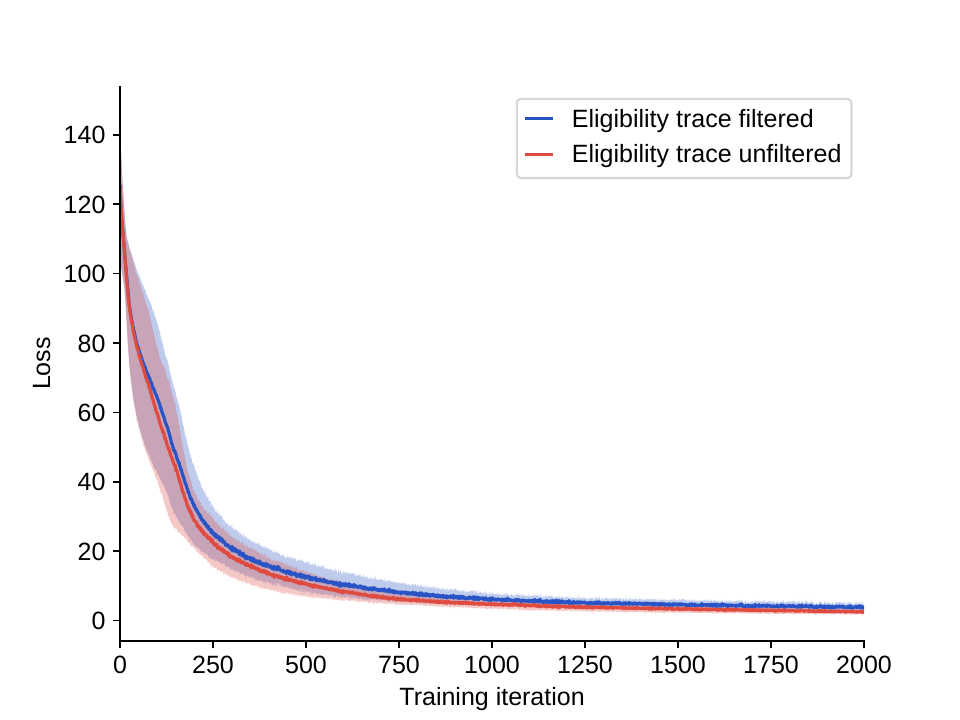}
      \caption[Learning performance comparison between models with filtered and unfiltered eligibility traces]{\captiontitle{Learning performance comparison between models with filtered and unfiltered eligibility traces.}
        Loss time courses for the pattern generation task simulated for \SI{2000}{} iterations and averaged over \SI{10}{} random seeds.
      }\label{fig:etrace-filter}
    \end{suppfigure}

  \section{Generalized delays}\label{sec:generalized-delays}

    The original model \supercite{bellec2020solution} assumes certain signal transmissions to be instantaneous, while the remaining delays are coupled to the temporal resolution of the simulation, which aligns with historical methods\supercite{lansner2012virtues}, as discussed in \cref{sec:transmission-delays}. Here, we introduce a more explicit and biologically motivated approach: we decouple these delays from the temporal resolution of the simulation and include explicit transmission times for all connections the original model considers instantaneous.

    In particular, the original model does not incorporate a delay between the occurrence of synaptic activities and the arrival of the learning signals that convey feedback about the network's output error. Accounting for this delay --- specifically, the time it takes for signals to travel from the recurrent layer to the output layer and back as error-related feedback --- is essential for addressing what is known as the `distal reward problem' \supercite{izhikevich2007solving,potjans2011imperfect}. Although this problem is usually framed in the context of reinforcement learning, it is also relevant to supervised learning scenarios. In these contexts, any feedback mechanism that modulates synaptic updates might reasonably be expected to exhibit some temporal lag.

    Delays may arise from several biophysical mechanisms. For instance, the propagation of learning signals may involve slower neuromodulatory processes, chemical intermediaries, or other non-instantaneous mechanisms that unfold over longer timescales \supercite{brzosko2019neuromodulation}. Additionally, neural circuits are spatially distributed \supercite{seguin2023brain}, and the physical distance between neural populations prolongs the travel time of a signal. By incorporating these transmission delays, we seek to more faithfully model the temporal structure of learning signals and thus improve the biological plausibility of the learning dynamics.

    \subsection{Delay from the recurrent to the output layer}\label{sec:recurrent-to-output-delay}

      Assuming a delay \(d\) in the transmission from the recurrent to the output layer leads to a delayed effect of the network activity \(z_j^t\) on the output state \(y_k^{t+d}\), and therefore also on the loss \(\loss \) (cf.\ \cref{eq:leaky_integrator,eq:mean-squared-error,eq:cross-entropy-loss}). To mathematically model this delay, we modify the equations for temporal credit assignment. Following an established notation \supercite{martin2022taxonomy}, we denote by \( \dvr[r]{}{\cdot} \) a ``readout-restricted derivative'', that is, a derivative which accounts for temporal recurrences within the readout layer while excluding any dependencies that propagate back into the recurrent network. Using this notation, the influence of \( z_j^t \) on \( \loss \), with \( \pdv{y_k^\tau}{z_j^t} = 0 \) for any \( \tau < t + d \), reads
      \begin{equation}
        \dvr[r]{\loss}{z^{t'}_j} = \sum_{k=1}^K \dvr[r]{\loss}{y^{t'+d}_k} \pdv{y^{t'+d}_k}{z^{t'}_j},
        \label{eq:readout-derivative-base}
      \end{equation}
      where
      \begin{align}
        \dvr[r]{\loss}{y^{t'+d}_k} = \pdv{\loss}{y^{t'+d}_k} + \dvr[r]{\loss}{y^{t'+d+1}_k} \pdv{y^{t'+d+1}_k}{y^{t'+d}_k}.
        \label{eq:dL_dy_base}
      \end{align}
      Expanding recursively:
      \begin{multline}
        \dvr[r]{\loss}{y^{t'+d}_k} = \pdv{\loss}{y^{t'+d}_k} + \pdv{\loss}{y^{t'+d+1}_k} \pdv{y^{t'+d+1}_k}{y^{t'+d}_k} + \pdv{\loss}{y^{t'+d+2}_k} \pdv{y^{t'+d+2}_k}{y^{t'+d+1}_k} \pdv{y^{t'+d+1}_k}{y^{t'+d}_k} \\
        + \pdv{\loss}{y^{t'+d+3}_k} \pdv{y^{t'+d+3}_k}{y^{t'+d+2}_k} \pdv{y^{t'+d+2}_k}{y^{t'+d+1}_k} \pdv{y^{t'+d+1}_k}{y^{t'+d}_k} + \cdots + \dvr[r]{\loss}{y^{T}_k} \pdv{y^T_k}{y^{T-1}_k} \cdots \pdv{y^{t'+d+1}_k}{y^{t'+d}_k}.
        \label{eq:dL_dy_recursion}
      \end{multline}
      However at the end of the sequence \( (t= T)\) the recursion stops and the readout-restricted derivative becomes equal to the partial derivative, i.e, \(\dvr[r]{\loss}{y^{T}_k} = \pdv{\loss}{y^{T}_k}\). Therefore, \cref{eq:dL_dy_recursion} can be compactly expressed as
      \begin{align}
        \dvr[r]{\loss}{y^{t'+d}_k} & = \sum^{T}_{t = t'+d} \pdv{\loss}{y^t_k} \pdv{y^t_k}{y^{t-1}_k} \cdots \pdv{y^{t'+d+1}_k}{y^{t'+d}_k}           \\
                                   & = \sum^{T}_{t = t'+d} \pdv{\loss}{y^t_k} \left( \prod^{t-1}_{\tau=t'+d} \pdv{y^{\tau+1}_k}{y^{\tau}_k} \right).
        \label{eq:dL_dy_recursion_compact}
      \end{align}
      Substituting \cref{eq:dL_dy_recursion_compact} back into \cref{eq:readout-derivative-base} yields
      \begin{align}
        \dvr[r]{\loss}{{z_j^{t'}}}
        = \sum_{k=1}^K \sum_{t=t'+d}^T \pdv{\loss}{y_k^t} \qty( \prod_{\tau=t'+d}^{t-1} \pdv{y_k^{\tau+1}}{y_k^\tau} ) \pdv{y_k^{t'+d}}{z_j^t}.
        \label{eq:dLdz-modified2}
      \end{align}

      Next, we compute the gradient of the loss function with respect to the recurrent weights:
      \begin{align}
        \dv{\loss}{W_{ji}^\text{r}} \approx \sum_{t=1}^T \dvr[r]{\loss}{{z_j^t}} e_{ji}^t.
        \label{eq:eprop-approx-final}
      \end{align}
      Substituting \cref{eq:eligibility-trace,eq:dLdz-modified2} into \cref{eq:eprop-approx-final}, we obtain
      \begin{align}
        \dv{\loss}{W_{ji}^\text{r}} \approx \sum_{k=1}^K \sum_{t'=1}^{T-d} \sum_{t=t'+d}^T \pdv{\loss}{y_k^t} \qty( \prod_{\tau=t'+d}^{t-1} \pdv{y_k^{\tau+1}}{y_k^\tau} ) \pdv{y_k^{t'+d}}{z_j^{t'}} \pdv{z_j^{t'}}{\vb*{h}_j^{t'}} \epsilon_{ji}^{t'}.
        \label{eq:eprop-approx-final-expanded}
      \end{align}
      Similarly, for the output weights, we derive
      \begin{align}
        \dv{\loss}{W_{kj}^\text{o}} & = \sum_{t=1}^T \pdv{\loss}{y_k^t} \dv{y_k^t}{W_{kj}^\text{o}}                                                                                \\
                                    & = \sum_{t'=1}^T \sum_{t=t'}^T \pdv{\loss}{y_k^t} \qty( \prod_{\tau=t'}^{t-1} \pdv{y_k^{\tau+1}}{y_k^\tau} ) \pdv{y_k^{t'}}{W_{kj}^\text{o}}.
        \label{eq:eprop-output}
      \end{align}
      To ensure the gradients do not depend on future errors, thereby maintaining correct temporal directionality in gradient computations, we invert the indices \(t'\) and \(t\) similar to the identity in \cref{eq:index-flipping} for an arbitrary matrix \(A\)
      \begin{align}
        \sum_{t'=1}^{T-d} \sum_{t=t'+d}^T A^{t',t} = \sum_{t=1+d}^T \sum_{t'=1}^{t-d} A^{t',t}.
        \label{eq:index-flipping-delay}
      \end{align}
      and we obtain
      \begin{align}
        \dv{\loss}{W_{ji}^\text{r}} & \approx \sum_{k=1}^K \sum_{t=1+d}^T \pdv{\loss}{y_k^t} \sum_{t'=1}^{t-d} \qty( \prod_{\tau=t'+d}^{t-1} \pdv{y_k^{\tau+1}}{y_k^\tau} ) \pdv{y_k^{t'+d}}{z_j^{t'}} \pdv{z_j^{t'}}{\vb*{h}_j^{t'}} \epsilon_{ji}^{t'},
        \label{eq:dL_dWrec_with_identity}                                                                                                                                                                                                                 \\
        \dv{\loss}{W_{kj}^\text{o}} & = \sum_{t=1}^T \pdv{\loss}{y_k^t} \sum_{t'=1}^t \qty( \prod_{\tau=t'}^{t-1} \pdv{y_k^{\tau+1}}{y_k^\tau} ) \pdv{y_k^{t'}}{W_{kj}^\text{o}}.
        \label{eq:eprop-output-new}
      \end{align}
      The second sum indexed by \(t'\) is now over previous events and can therefore be computed online. These backward-looking expressions allow for cumulative computation, avoiding the need to defer calculations until all future errors are known.

      By incorporating the output delay into the dynamics of the output neurons in \cref{eq:leaky_integrator}, we get
      \begin{align}
        y_k^t = \kappa y_k^{t-1} + \sum_{j=1}^J W_{kj}^\text{o} z_j^{t-d},
        \label{eq:leaky_integrator_with_delay}
      \end{align}
      from which we can calculate the derivatives
      \begin{align}
        \pdv{y_k^{t+d}}{z_j^t}       & = W_{kj}^\text{o}, \\
        \pdv{y_k^t}{W_{kj}^\text{o}} & = z_j^{t-d},       \\
        \pdv{y_k^{\tau+1}}{y_k^\tau} & = \kappa.
      \end{align}
      This allows us to simplify the product in \cref{eq:dL_dWrec_with_identity}:
      \begin{align}
        \prod_{\tau=t'+d}^{t-1} \pdv{y_k^{\tau+1}}{y_k^\tau} = \clash{\kappa}{t-d-t'}.
      \end{align}
      using \cref{not:exponent-notation}.
      By substituting this expression into \cref{eq:dL_dWrec_with_identity}, we get
      \begin{align}
        \dv{\loss}{W_{ji}^\text{r}} & \approx \sum_{k=1}^K \sum_{t=1+d}^T W_{kj}^\text{o} E_k^t \sum_{t'=1}^{t-d} \clash{\kappa}{t-d-t'} \psi_j^{t'} \filter{F}{\alpha}{z_i^{t'-1}} \\
                                    & = \sum_{t=1+d}^T L_j^t \filter{F}{\kappa}{\psi_j^{t-d} \filter{F}{\alpha}{z_i^{t-d-1}}},
        \label{eq:eprop-approx-final-expanded-subs-rewritten}
      \end{align}
      and by substituting it into \cref{eq:eprop-output-new}
      \begin{align}
        \dv{\loss}{W_{kj}^\text{o}} & = \sum_{t=1+d}^T E_k^t \sum_{t'=1}^t \clash{\kappa}{t-t'} z_j^{t'-d} \\
                                    & = \sum_{t=1+d}^T E_k^t \filter{F}{\kappa}{z_j^{t-d}}.
        \label{eq:eprop-output-new-subs-rewritten}
      \end{align}
      The learning signal generated at time \(t\) must be paired with the eligibility trace value at \(t - d\), as the network activity prior to the output delay caused the corresponding error.

    \subsection{Delay from the output to the recurrent layer}\label{sec:output-to-recurrent-delay}

      Here we consider the case where there is a non-zero delay \( d^\text{ls} \) in the transmission of the learning signal from the output layer to the recurrent layer. This means that at step \( t \), the most recent learning signal available at a recurrent neuron \( j \) is \( L_j^{t - d^\text{ls}} \) and the most recent instantaneous gradient that can be applied is \(g_{ji}^{t -d^\text{ls}}\). If the transmission delay from the recurrent to the output layer is \( d \), we write the learning rule for the recurrent synapses as
      \begin{align}
        \pdv{\loss}{W_{ji}^\text{r}} & = \sum_{t=1+d+d^\text{ls}}^T g_{ji}^{t-d^\text{ls}}                                                                                        \\
                                     & = \sum_{t=1+d+d^\text{ls}}^T L_j^{t-d^\text{ls}} \filter{F}{\kappa}{\psi_j^{t-d-d^\text{ls}} \filter{F}{\alpha}{z_i^{t-d-d^\text{ls}-1}}}.
        \label{eq:eprop-learning-signal-delay}
      \end{align}
      Thus, the instantaneous gradient contribution that would normally be computed at time \(t\) is instead computed only when the learning signal arrives in the network at the later time \(t + d^{\text{ls}}\), and the weight updates are based on these delayed gradients. In the case of gradient descent, the resulting scheme is known as delayed gradient descent (DGD), widely used in distributed or asynchronous optimization scenarios \supercite{aviv2021asynchronous}. Asynchronous delays in gradient descent have been shown to reduce generalization error \supercite{deng2025toward}. Since the learning signal is generated at the output layer, the output layer weights do not depend on the learning signal delay. Therefore, the learning rule for the output synapses remains unchanged.

      We propose two hypotheses regarding how evolution might have fine-tuned brain circuits to handle these delays in credit assignment. First, the time required to convert the surrogate gradient and presynaptic spikes into an eligibility trace could have been synchronized by evolution to match the time it takes for signals to travel from the recurrent to the output layer and back. This synchronization delay can be mathematically expressed as
      \begin{align}
        e_{ji}^{\text{sync},t} = \psi_j^{t-d^\text{sync}} \filter{F}{\alpha}{z_i^{t-d^\text{sync}-1}}.
      \end{align}
      Second, neural circuits may have evolved to process error feedback with a delay, reflecting the time required for signals to arrive from higher brain areas.

  \section{Algorithms}\label{sec:algorithms}

    The time-driven algorithm for the computation of the gradient (see \cref{alg:time-driven-gradient-computation}) updates the eligibility trace in each step, but the new value contributes to the gradient only after a delay of \(d\) steps. To manage this delay, we use a first-in-first-out (FIFO) queue (\cref{fig:fifo-queue}).

    \begin{suppfigure}[H]
      \centering
      \begin{tikzpicture}[square/.style={draw, minimum width=1.5cm, minimum height=1cm, inner sep=0pt}, >=Stealth]
        \node         (V1) at (0.0,1.0) {\centering \( e_{ji}^{t-d} \)};
        \node[square] (V2) at (1.5,0.0) {\centering \( e_{ji}^{t-d+1} \)};
        \node[square] (V3) at (3.5,0.0) {\centering \( e_{ji}^{t-d+2} \)};
        \node         (V4) at (4.8,0.0) {\centering \( \cdots \)};
        \node[square] (V5) at (6.0,0.0) {\centering \( e_{ji}^{t-2} \)};
        \node[square] (V6) at (8.0,0.0) {\centering \( e_{ji}^{t-1} \)};
        \node         (V7) at (9.5,1.0) {\centering \( e_{ji}^t \)};

        \draw[->] (V2.west)  to[bend left] (V1.south);
        \draw[->] (V7.south) to[bend left] (V6.east);
        \draw[->] (V3.south) to[bend left] (V2.south);
        \draw[->] (V6.south) to[bend left] (V5.south);

      \end{tikzpicture}
      \caption[Eligibility trace buffering with FIFO queue for weight updates with every spike.]{Eligibility trace buffering using a FIFO queue for weight updates with every spike. Each box in the FIFO queue represents the eligibility trace value at a specific step. The newest entry is enqueued, while the oldest is dequeued and used to compute the gradient.
      }\label{fig:fifo-queue}
    \end{suppfigure}

    Since an input spike has an effect on the recurrent layer earliest after a travel time of 1 step at \( t=1 \), we can assume for any \( t < 1 \) that \( \psi_j^t\), \( z_i^t \), \( f_j^t \), and thus \( e_{ji}^t \) are zero and initialize the eligibility trace queue in \cref{fig:fifo-queue} with zeros. With this initialization, the summations in \cref{eq:eprop-approx-final-expanded-subs-rewritten,eq:eprop-output-new-subs-rewritten} can start from \(t=1\) instead of \(t=1+d\), effectively simplifying the initial conditions for the procedures outlined in \cref{alg:time-driven-gradient-computation}. An alternative approach is to apply the weight updates immediately at each step, as shown in \cref{alg:time-driven-weight-update}. Analogous to the buffering of state variables for recurrent synapses, the algorithm updates output synapses by buffering \(z_i^t\). The pseudocode \cref{alg:event-driven-weight-update} illustrates the event-driven weight update.

    When performing event-driven updates, a particular simplification becomes feasible. For recurrent synapses, the postsynaptic neuron maintains a history of surrogate gradients. Consequently, it is sufficient to maintain a FIFO queue for the incoming presynaptic spike state variable at each step \( t \), denoted by \( z_i^{t-1} \) (see \cref{alg:event-driven-weight-update-new}). Since \( z_i^{t-1} \) is a binary variable, it is sufficient to only store the timestamps \( t \) in the interval \( [t - d^\text{sync}, t ] \), when \( z_i^{t-1} = 1 \). This selective storage enables the algorithm to compute the eligibility trace on demand using only the relevant spike times, thereby reducing memory requirements. The idea of recording only the relevant spike times readily transfers to output synapses. The pseudocode \cref{alg:event-driven-weight-update-optimized} illustrates the optimized event-driven weight update.

    % chktex-file 1
    \begin{algorithm}[H]
      \begin{algorithmic}[1]
        \Require \(W_{ji}^\text{r}, T, d, f\qty(\cdot)\)
        \State{Initialize gradient sum \(g_{ji} \gets 0\)}
        \For{\(t = 1\) to \(T\)}
        \State Enqueue current eligibility trace \( e_{ji}^t = \psi_j^t \filter{F}{\alpha}{z_i^{t-1}} \)
        \State Dequeue oldest eligibility trace \( e_{ji}^{t-d} \)
        \State Update gradient sum
        \State \(g_{ji} \gets g_{ji} + L_j^t \filter{F}{\kappa}{e_{ji}^{t-d}} + c^{\text{reg},*}\qty(f_j^{t-d} - f^*) \filter{G}{\beta}{e_{ji}^{t-d}} \)
        \EndFor
        \State Update weight \(W_{ji}^\text{r} \gets W^\text{r}_{ji} + f\qty(g_{ji})\)
        \State \Return \(W_{ji}^\text{r}\)
      \end{algorithmic}
      \caption{Time-driven gradient computation}\label{alg:time-driven-gradient-computation}
    \end{algorithm}

    \begin{algorithm}[H]
      \begin{algorithmic}[1]
        \Require \(W_{ji}^{\text{r}}, T, d, f\qty(\cdot)\)
        \For{\(t = 1\) to \(T\)}
        \State Enqueue current eligibility trace \(e_{ji}^t = \psi_j^t \filter{F}{\alpha}{z_i^{t-1}} \)
        \State Dequeue oldest eligibility trace \(e_{ji}^{t-d}\)
        \State Compute gradient contribution
        \State \(g_{ji} \gets L_j^t \filter{F}{\kappa}{e_{ji}^{t-d}} + c^{\text{reg},*} \qty(f_j^{t-d} - f^*) \filter{G}{\beta}{e_{ji}^{t-d}}\)
        \State Update weight \(W_{ji}^\text{r} \gets W_{ji}^\text{r} + f\qty(g_{ji})\)
        \EndFor
        \State \Return updated \(W_{ji}^\text{r}\)
      \end{algorithmic}
      \caption{Time-driven weight update}\label{alg:time-driven-weight-update}
    \end{algorithm}

    \begin{algorithm}[H]
      \begin{algorithmic}[1]
        \Require \(W_{ji}^{\text{r}}, t^\text{prev spike}, t^\text{spike}, d, f\qty(\cdot)\)
        \For{\(t = t^\text{prev spike}\) to \(t^\text{spike}-1\)}
        \State Enqueue current eligibility trace \(e_{ji}^t = \psi_j^t \filter{F}{\alpha}{z_i^{t-1}}\)
        \State Dequeue oldest eligibility trace \(e_{ji}^{t-d}\)
        \State Compute gradient contribution \(g_{ji} \gets L_j^t \filter{F}{\kappa}{e_{ji}^{t-d}}\)
        \State Update weight \(W_{ji}^\text{r} \gets W_{ji}^\text{r} + f\qty(g_{ji})\)
        \EndFor
        \State \Return updated \(W_{ji}^\text{r}\)
      \end{algorithmic}
      \caption{Event-driven weight update}\label{alg:event-driven-weight-update}
    \end{algorithm}

    \begin{algorithm}[H]
      \begin{algorithmic}[1]
        \Require \(W_{ji}^{\text{r}}, t^\text{prev spike}, t^\text{spike}, d, f\qty(\cdot)\)
        \For{\(t = t^\text{prev spike} + 1\) to \(t^\text{spike}\)}
        \State Enqueue current incoming presynaptic spiking state \(z_i^{t-1}\)
        \State Dequeue oldest presynaptic spiking state \(z_i^{t-d-1}\)
        \State Compute \(e_{ji}^{t-d} = \psi_j^{t-d} \filter{F}{\alpha}{z_i^{t-d-1}}\)
        \State Compute gradient contribution
        \State \(g_{ji} \gets L_j^t \filter{F}{\kappa}{e_{ji}^{t-d}} + c^{\text{reg},*} \qty(f_j^{t-d} - f^*) \filter{G}{\beta}{e_{ji}^{t-d}}\)
        \State Update weight \(W_{ji}^\text{r} \gets W_{ji}^\text{r} + f\qty(g_{ji})\)
        \EndFor
        \State \Return updated \(W_{ji}^\text{r}\)
      \end{algorithmic}
      \caption{Event-driven weight update}\label{alg:event-driven-weight-update-new}
    \end{algorithm}

    \begin{algorithm}[H]
      \begin{algorithmic}[1]
        \Require \(W_{ji}^\text{r}, t^\text{prev spike}, t^\text{spike}, d, f\qty(\cdot)\)
        \For{\(t = t^\text{prev spike} + 1\) to \(t^\text{spike}\)}
        \State Store \(t\) in \(\text{times}\) if \(z_i^{t-1} = 1\)
        \If{\(t - d\) is in \(\text{times}\)}
        \State Compute \(e_{ji}^{t-d} \gets \psi_j^{t-d} \filter{F}{\alpha}{1}\)
        \State Remove \(t-d\) from \(\text{times}\)
        \Else
        \State Compute \(e_{ji}^{t-d} \gets \psi_j^{t-d} \filter{F}{\alpha}{0}\)
        \EndIf
        \State Compute gradient contribution
        \State \(g_{ji} \gets L_j^t \filter{F}{\kappa}{e_{ji}^{t-d}} + c^{\text{reg},*} \qty(f_j^{t-d} - f^*) \filter{G}{\beta}{e^{t-d}_{ji}}\)
        \State Update weight \(W_{ji}^\text{r} \gets W_{ji}^\text{r} + f\qty(g_{ji})\)
        \EndFor
        \State{\Return{updated \(W_{ji}^\text{r}\)}}
      \end{algorithmic}
      \caption{Optimized event-driven weight update}\label{alg:event-driven-weight-update-optimized}
    \end{algorithm}
    % chktex-file +1

  \section{Architecture}\label{sec:architecture}
    This section presents the software architecture underlying our reference implementation of the e-prop neuron and synapse models. To ensure compatibility with the modular design \citep{brette2007simulation} of the reference code, we decompose the mathematical models into objects that represent neurons with e-prop histories, synapses that retrieve those histories, and connectivity components that deliver learning signals.

    When a presynaptic spike triggers a weight update, the event-driven scheme accesses the postsynaptic histories of several quantities between the last update and the current time, accounting for dendritic delays. For recurrent neurons, the relevant quantities are the surrogate gradients, firing-rate regularization, and learning signals, whereas for output neurons they are the error signals.

    In hybrid parallelization on a cluster of compute nodes, where OpenMP threads manage intra-node parallelization, while the Message Passing Interface (MPI) handles inter-node communication, it is computationally efficient to allocate synapses on the same MPI process as their postsynaptic neurons \citep{morrison2005advancing}. In this setup, no inter-process communication is required to retrieve postsynaptic information for weight updates, and each source neuron communicates a single spike to all its target neurons on a given MPI process.

    A key precedent for our work is the \code{ArchivingNode} class \supercite{morrison2007spike}. This class serves as a parent for neuron models supporting STDP in event-driven simulations by providing member functions to store and retrieve postsynaptic spike histories. Incoming synapses access this archived information to perform event-driven weight updates. Building on this concept, the challenge of implementing voltage-based plasticity rules --- where synaptic weight changes depend continuously on postsynaptic membrane voltages --- was tackled previously within an event-driven framework \supercite{stapmanns2021event}. The authors align two prominent plasticity models \supercite{clopath2010connectivity,urbanczik2014learning} in a common framework and introduce the corresponding re-implementations \code{ClopathArchivingNode} and \code{UrbanczikArchivingNode}, respectively. Their approach extends the parent classes with a vector-based archive that stores information derived from the membrane voltage. When an incoming spike occurs, synapses use this stored data for plasticity computations, enabling an event-driven implementation of these rules.

    Following this line of work, we introduce the new class \code{EpropArchivingNode}, which provides a common interface that maintains a vector-based archive of e-prop signals. We further define specialized subclasses that expose methods to store and retrieve entries from these archives. These subclasses are neuron-specific, as the incoming synapses to different neuron types require distinct data.\ \code{EpropArchivingNodeReadout} handles the history of error signals for output neurons, whereas \code{EpropArchivingNodeRecurrent} manages the histories of surrogate gradients, learning signals, and firing-rate regularization for recurrent neurons.

    Objects representing neuron models that support e-prop inherit indirectly from \code{EpropArchivingNode} through the corresponding specialized subclass and employ the subclass's methods to record new data. To differentiate models implementing the event-driven variant of the original e-prop algorithm from those incorporating further biological constraints, we append the suffix \code{bsshslm_2020} to the former. We derive this suffix from the initials of the authors of the original publication \supercite{bellec2020solution} and the year of publication. In \code{bsshslm_2020} models, synaptic updates occur only after processing an entire training sample, with the \code{update_interval} set to the sample duration. Our framework defines update times as \code{t_update} = \code{shift} + \code{update_interval} \(\times \; i\), where \( i=0,1,\dots \). The \code{shift} parameter aligns weight updates to reproduce the time-driven results \supercite{bellec2020solution}. Specifically, the \code{shift} value represents the minimum number of steps required for an input spike to reach the recurrent layer (for recurrent neurons) or the output layer (for output neurons). A difference of 1 in the \code{shift} value between recurrent and output neurons accounts for the extra step spikes require to reach the output layer. In the time-driven implementation, the output layer integrates recurrent spikes without delay; thus, the additional shift for output neurons ensures alignment with those results. In \epropplus{}, there is no need to replicate the time-driven results, nor is a relative shift between recurrent and output neurons required, as the delay between these layers is already embedded in the learning rule. Moreover, each new spike immediately triggers a weight update, so that \code{t_update} = \code{t_spike}.

    In both cases, recurrent neurons call a method, \code{append_new_eprop_history_entry}, at each step to insert an empty history entry. This entry is then populated with surrogate gradients and learning signals using the methods \sloppy{\code{write_surrogate_gradient_to_history}} and \code{write_learning_signal_to_history}, respectively. Output neurons follow a similar procedure to record their error signals.

    Since the exact plasticity rule depends on the dynamics of the postsynaptic neuron, we implement the gradient computation as a member function of each neuron model named \code{compute_gradient}. Incoming synapses invoke this function when a presynaptic spike triggers a weight update, passing all presynaptic spike times since the last such update. In \epropplus{}, each new spike triggers a weight update, which significantly reduces the amount of spike data that must be transmitted. However, in this scenario, the synapse also forwards the values of the relevant running synaptic traces, as they are not reset between updates. The neuron object then computes the new gradients required for the weight update using the provided spike times --- and, if necessary, the synaptic trace values --- and returns the gradient to the synapse for the update. The synapse objects with e-prop functionality derive from a standard \code{Connection} class following the same naming convention (adding the suffix \code{bsshslm_2020} to the base version name).

  \section{History Management}\label{sec:history-management}

    To optimize memory usage and prevent an unnecessary growth of \code{eprop_history} for each neuron, a cleaning mechanism removes entries that have already been utilized by the weight updates of all incoming synapses. This mechanism is based on a second vector-based archive, \code{update_history}, which each postsynaptic neuron locally maintains. The \code{update_history} is a vector of pairs \code{update_history} = (\code{t_update}, \code{access_counter}), where \code{t_update} indicates the time point at which a synapse's weight is updated, and \code{access_counter} records the number of synapses whose most recent update occurred at that time. When an incoming synapse triggers a weight update at \code{t_update} \(= t_1\) and had a previous update at \code{t_update} \(= t_0\), the \code{access_counter} for \(t_0\) is decreased by 1, and if it reaches 0, the entry corresponding to \(t_0\) is removed from \code{update_history}. Simultaneously, if an entry for \(t_1\) already exists, its \code{access_counter} is incremented by 1; if not, a new entry is created with an \code{access_counter} set to 1. Consequently, the growth of \code{update_history} is inherently bounded by the number of synapses targeting the neuron. Using this auxiliary history, which monitors the progress of weight updates for all incoming synapses, the algorithm identifies sections of the \code{eprop_history} that are no longer required.

    Since \epropplus{} does not require samples of fixed length --- unlike the base event-driven models (denoted by the suffix \code{bsshslm_2020}) --- we implement a slightly different cleaning strategy for each case. In the \code{bsshslm_2020} version, each e-prop neuron calls the function \code{erase_used_eprop_history} at the beginning of every sample. This function removes both outdated and redundant \code{eprop_history} entries. It eliminates outdated entries by deleting all history records that precede the \code{t_update} entry at the front of the \code{update_history}, and retains all other entries until they have been used by all incoming synapses. However, neurons often remain silent for extended periods and may not emit any spikes throughout an entire sample. In such cases, the gradients for their outgoing connections remain zero, making it unnecessary to compute or store their history entries. To exploit this, all \code{eprop_history} entries corresponding to samples that lack a matching \code{update_history} entry (i.e., those with an \code{access_counter} of zero) are safely removed.

    In \epropplus{}, which accumulates the gradients for a weight update between spikes, each neuron calls the function \code{erase_used_eprop_history} for every incoming spike, rather than once at the beginning of a sample as in \code{bsshslm_2020}. The \epropplus{} algorithm eliminates outdated entries by deleting all history entries older than \code{t_update} at the front of the \code{update_history}. Also in this version, silent neurons lead to uncontrolled growth of \code{eprop_history}. Since the fixed-length updates are absent, we cannot employ the scheme of the \code{bsshslm_2020} models to deal with this circumstance. Instead, we introduce a cutoff to limit e-prop history accumulation between spikes. The rapid decay of the eligibility trace justifies this cutoff, making it a practical approximation in sparse spiking scenarios. Because all entries in \code{update_history} are sorted by \code{t_update}, identifying removable entries is straightforward: if two consecutive entries in \code{update_history} have \code{t_update} values \(t_0\) and \(t_1\), then the algorithm removes every \code{eprop_history} entry between \(t_0~+\)~\code{eprop_isi_trace_cutoff} and \(t_1\), as these entries lie within the cutoff region and will not be used by any incoming synapse.

  \section{Locality, causality, and onlineness}\label{sec:locality-causality-onlineness}
    According to a recent framework \supercite{martin2022taxonomy}, e-prop can be viewed as a local and causal rule and as an approximation of the non-local, non-causal backpropagation through time (BPTT) and non-local Real-Time Recurrent Learning (RTRL), both of which compute exact gradients. RTRL is computationally more intensive because it maintains eligibility traces that are updated recursively at each step to ensure causality \supercite{martin2022taxonomy}. Here, non-causal refers to requiring knowledge of future activity, while non-local implies that the error must be propagated across all neurons and synapses \supercite{martin2022taxonomy}. Unlike BPTT, e-prop and RTRL are classified as online gradient computation algorithms, as their computations for a given step are causal and can be performed during the forward pass \supercite{martin2022taxonomy_arxiv}.

    In e-prop and in the typical definition of BPTT, weights remain constant over the mini-batch. Under this condition, BPTT is equivalent to RTRL\@. An online weight update algorithm, in contrast, updates weights continuously, which produces results distinct from those of these three algorithms \supercite{martin2022taxonomy_arxiv}.

    Truncated algorithms update weights more frequently than once per mini-batch, enabling short-term pattern learning and faster adaptation to input changes, but reducing the ability to capture long-term dependencies and lowering accuracy by approximating full-sequence gradients. In these algorithms, gradients are computed with outdated parameters, specifically synaptic weights. This trade-off between dynamic adaptation and gradient accuracy \supercite{menick2020practical} must be balanced against task requirements. Weights may be updated midway through a sample \supercite{budik2006trtrl} or at every step \supercite{kag2021training}; the latter was also demonstrated in SNNs \supercite{yin2023accurate}.

  \section{Online training algorithms of recurrent neural networks}\label{sec:online-training-recurrent-nets}

    This section provides an overview of online training algorithms for biologically plausible recurrent neural networks and efforts to systematize them. A recent framework \supercite{marschall2020unified} organizes state-of-the-art algorithmic developments that are efficient, biologically plausible, or both, along criteria such as past- vs.\ future-facing, tensor structure, stochastic vs.\ deterministic, and closed-form vs.\ numerical solutions. The authors provide mathematical intuitions for their success and compare BPTT and RTRL with Unbiased Online Recurrent Optimization (UORO) \supercite{tallec2017unbiased}, Kernel RNN Learning (KeRNL) \supercite{roth2018kernel}, Kronecker-Factored RTRL (KF-RTRL) \supercite{mujika2018approximating}, RFLO \supercite{murray2019local}, r-Optimal Kronecker-Sum Approximation (r-OK) \supercite{benzing2019optimal}, and Decoupled Neural Interfaces (DNI) \supercite{jaderberg2017decoupled}, noting that e-prop is essentially equivalent to RFLO\@. They also propose new variants, including Reverse KF-RTRL (R-KF), Efficient BPTT (E-BPTT), and Future-Facing BPTT (F-BPTT). Other algorithms outside this framework include a spiking backpropagation variant \supercite{lee2016training}, Sparse n-step Approximation (SnAP) \supercite{menick2020practical}, Deep Continuous Local Learning (DECOLLE) \supercite{kaiser2020synaptic}, Online Spatio-Temporal Learning (OSTL) \supercite{bohnstingl2022online}, Forward Propagation (FP) \supercite{lee2022exact}, Event-based Three-Factor Local Plasticity (ETLP) \supercite{quintana2024etlp}, as well as Spike-Timing-Dependent Event-Driven (STD-ED) and Membrane-Potential-Dependent (MP-ED) algorithms \supercite{wei2024event} and Smooth Exact Gradient Descent \supercite{klos2025smooth}, which addresses instability from spike appearance or disappearance described in \cref{sec:tasks}. Another rule \supercite{liu2021cell} enhances biological plausibility by incorporating neuron-type diversity and type-specific local neuromodulation.

    \printbibliography%
\end{document}